\documentclass[sigconf]{acmart}  

\usepackage{mathtools}
\usepackage{amsmath}
\usepackage{bm}
\usepackage{amsfonts}
\usepackage{amsbsy}
\usepackage{mathrsfs}
\usepackage[mathscr]{eucal}

\usepackage{hyperref}
\usepackage{subfigure}
\usepackage[normalem]{ulem}

\usepackage{algorithm}
\usepackage{algorithmic}
\usepackage{color}
\usepackage{makecell}
\usepackage{xspace}
\usepackage{bm}

\usepackage{multicol}
\usepackage{multirow}
\usepackage{mdwlist}
\usepackage{paralist}
\usepackage{enumerate}
\usepackage{breakcites}

\newcommand{\predmodel}{$\mathcal{M}$\xspace}
\newcommand{\ttq}{TDC\xspace}
\newcommand{\tsqfull}{temporal distribution characterization\xspace}
\newcommand{\cea}{TDM\xspace}

\newcommand{\tcs}{Temporal Covariate Shift\xspace}
\newcommand{\dtw}{Dynamic time wrapping\xspace}

\newtheorem{definition}{Definition}

\newtheorem{problem}{Problem}

\newtheorem{remark}{Remark}

\newcommand{\method}{\textsc{AdaRNN}\xspace}

\copyrightyear{2021}
\acmYear{2021}
\setcopyright{acmcopyright}
\acmConference[CIKM '21] {Proceedings of the 30th ACM International Conference on Information and Knowledge Management}{November 1--5, 2021}{Virtual Event, Australia.}
\acmBooktitle{Proceedings of the 30th ACM Int'l Conf. on Information and Knowledge Management (CIKM '21), November 1--5, 2021, Virtual Event, Australia}
\acmPrice{15.00}
\acmISBN{978-1-4503-8446-9/21/11}
\acmDOI{10.1145/XXXXXX.XXXXXX}

\begin{document}

\title{\method{}: Adaptive Learning and Forecasting for Time Series}
\titlenote{Corresponding author: J. Wang and W. Feng. This work is done when Y. Du was an intern at MSRA.}
\author{Yuntao Du$^1$, Jindong Wang$^2$, Wenjie Feng$^3$, Sinno Pan$^4$, Tao Qin$^2$, Renjun Xu$^5$, Chongjun Wang$^1$}
\affiliation{%
  \institution{$^1$Nanjing University, Nanjing, China \quad $^2$Microsoft Research Asia, Beijing, China}
  \city{$^3$Institute of Data Science, National University of Singapore \, $^4$Nanyang Technological University \, $^5$Zhejiang University}
\country{}
}
\email{dz1833005@smail.nju.edu.cn,  jindong.wang@microsoft.com}
\renewcommand{\shortauthors}{Du, et al.}










\begin{abstract}
Though time series forecasting has a wide range of real-world applications, it is a very challenging task. 
This is because the statistical properties of a time series can vary with time, 
causing the distribution to change temporally, which is known as the distribution shift problem in the machine learning community. 
By far, it still remains unexplored to model time series from the distribution-shift perspective. 
In this paper, we formulate the \tcs (TCS) problem for the time series forecasting. 
We propose Adaptive RNNs (\method) to tackle the TCS problem. 
\method is sequentially composed of two modules. The first module is referred to as 
Temporal Distribution Characterization, which aims to better 
characterize the distribution information in a time series. 
The second module is termed as Temporal Distribution Matching, 
which aims to reduce the distribution mismatch in the time series to 
learn an RNN-based adaptive time series prediction model. 
\method is a general framework with flexible distribution distances integrated. 
Experiments on human activity recognition, air quality prediction, 
household power consumption, and financial analysis show that 
\method outperforms some state-of-the-art methods by $2.6\%$ in terms of accuracy 
on classification tasks and by $9.0\%$ in terms of the mean squared error on regression tasks. 
We also show that the temporal distribution matching module can be extended to the 
Transformer architecture to further boost its performance.
\end{abstract}

\ccsdesc[500]{Computing methodologies~Transfer learning}
\ccsdesc[300]{Computing methodologies~Artificial intelligence}

\keywords{Time series, multi-task learning, transfer learning}

\maketitle
\section{Introduction}

Time series (TS) data occur naturally in countless domains including financial analysis~\cite{zhu2002statstream}, medical analysis~\cite{matsubara2014funnel},
weather condition prediction~\cite{vincent2019shape}, and renewable energy production~\cite{choi2016retain}.
Forecasting is one of the most sought-after tasks on analyzing time series data
(arguably the most difficult one as well) due to its importance in industrial,  society and scientific applications.
For instance, given the historical air quality data of a city for the last five days, 
how to predict the air quality in the future more accurately?

In real applications, it is natural that the statistical properties of TS are changing over time, i.e., the non-stationary TS.
Over the years, various research efforts have been made for building reliable and accurate models for the non-stationary TS.
Traditional approaches such as hidden Markov models (HMMs)~\cite{tuncel2018autoregressive}, 
dynamic Bayesian networks~\cite{robinson2010learning}, Kalman filters~\cite{de2020normalizing}, 
and other statistical models (e.g. ARIMA~\cite{kalpakis2001distance}), have made great progress.
Recently, better performance is achieved by the recurrent neural networks~(RNNs)~\cite{salinas2020deepar,vincent2019shape}.
RNNs make no assumptions on the temporal structure and can find highly non-linear and 
complex dependence relation in TS.


\begin{figure}[t!]
	\centering
	\includegraphics[width=.4\textwidth]{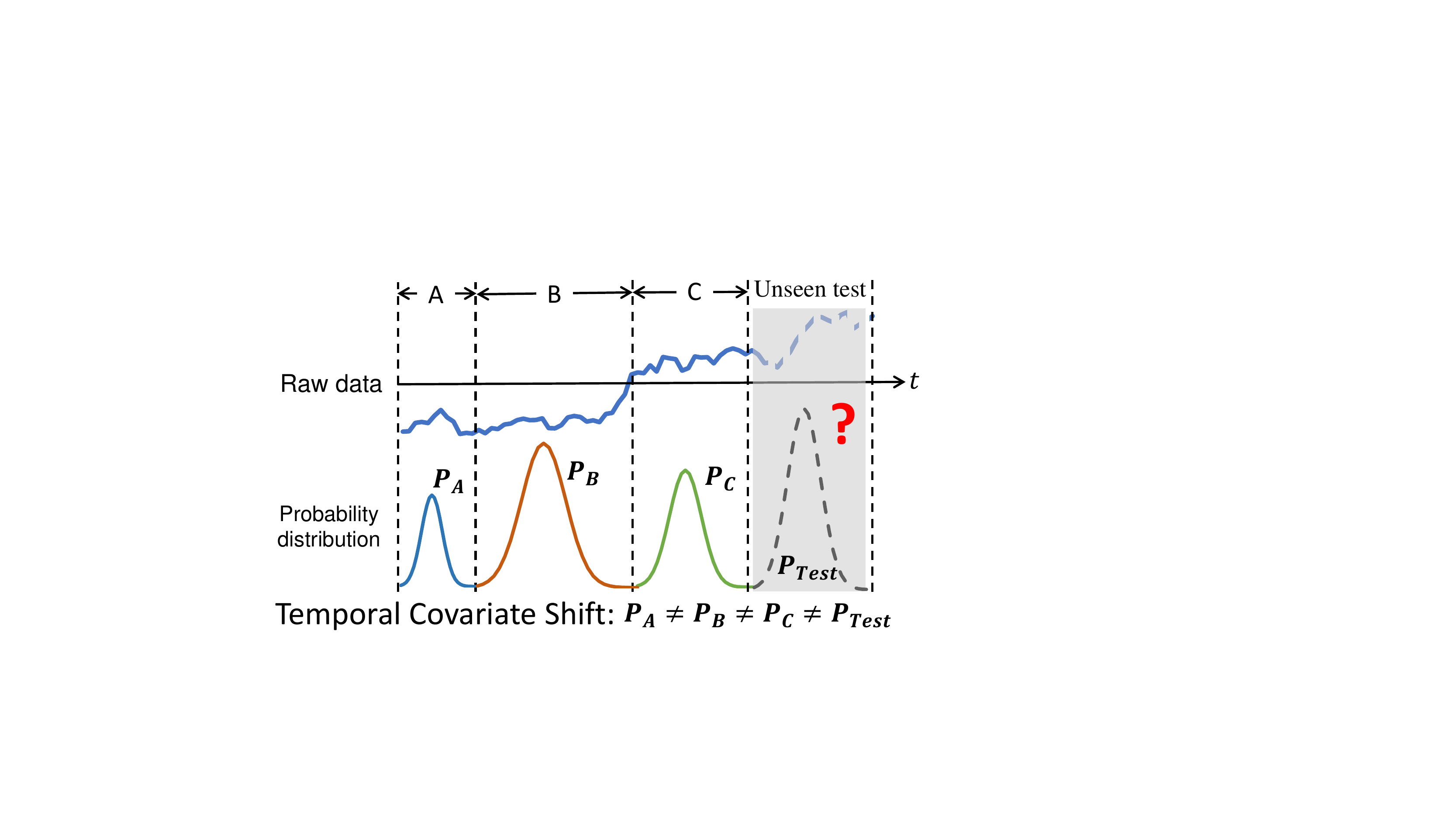}
	\caption{The \emph{temporal covariate shift} problem in non-stationary time series. The raw time series data ($x$) is multivariate in reality. At time intervals $A, B, C$ and the unseen test data, the distributions are different: $P_A(x) \ne P_B(x) \ne P_C(x) \ne P_{Test}(x)$. With the distribution changing over time, how to build an accurate and adaptive model?}
	\label{fig-motivation}
\end{figure}

The non-stationary property of time series implies the \emph{data distribution changes over time}. 
Given the example in \figurename~\ref{fig-motivation},
data distributions $P(x)$ vary for different intervals $A$, $B$, and $C$ 
where $x, y$ are samples and predictions respectively;
Especially for the test data which is unseen during training, 
its distribution is also different from the training data and makes the prediction more exacerbated.
The conditional distribution $P(y|x)$, however, is usually considered to be unchanged for this scenario, 
which is reasonable in many real applications.
For instance, in stock prediction, it is natural that the market is fluctuating 
which changes the financial factors ($P(x)$), while the economic laws remain unchanged ($P(y|x)$).
It is an inevitable result that the afore-mentioned methods have inferior performance and 
poor generalization~\cite{kuznetsov2014generalization} since the distribution shift issue 
violates their basic I.I.D assumption.


Unfortunately, it remains unexplored to model the time series from the distribution perspective.
The main challenges of the problem lies in two aspects.
First, 
how to characterize the distribution in the data to maximally harness the common knowledge in these varied distributions? 
Second, 
how to invent an RNN-based distribution matching algorithm to maximally reduce their distribution divergence while capturing the temporal dependency?

In this paper, we formally define 
the \emph{Temporal Covariate Shift (TCS)} in \figurename{~\ref{fig-motivation}}, which is a more practical and challenging setting to model time series data.
Based on our analysis on TCS, we propose \method, a novel framework 
to learn an accurate and adaptive prediction model. 
\method is composed of two modules.
Firstly, to better characterize the distribution information in TS, we propose a \textit{\tsqfull (TDC)} algorithm to split the training data into $K$ most diverse periods that are with large distribution gap inspired by the principle of maximum entropy.
After that, we propose a \textit{temporal distribution matching (TDM)} algorithm to dynamically reduce distribution divergence using a RNN-based model.
Experiments on activity recognition, air quality prediction, household power consumption and stock price prediction show that our \method outperforms the state-of-the-art baselines by $2.6\%$ in terms of an accuracy on classification tasks and by $9.0\%$ in terms of RMSE on regression tasks.
\method is agnostic to both RNN structures (i.e., RNNs, LSTMs, and GRUs) and distribution matching distances (e.g., cosine distance, MMD~\cite{borgwardt2006integrating}, or adversarial discrepancy~\cite{Ganin2016DomainAdversarialTO}).
\method can also be extended to the Transformer architecture to further boost its performance.

To sum up, our main contributions are as follows, 
\begin{itemize}
	\item{\bf{Novel problem:}} 
	    For the first time, we propose to model the time series from the distribution perspective, then we postulate and formulate the \tcs (TCS) problem in non-stationary time series, which is more realistic and challenging.
	\item{\bf{General framework:}} 
	    To solve TCS, we propose a general framework \method that learns an accurate and adaptive model by proposing the temporal distribution characterization and Temporal Distribution matching algorithms.
	\item{\bf{Effectiveness:}} 
	    We conduct extensive experiments on human activity recognition, air quality prediction, household power consumption, and stock price prediction datasets. 
	      Results show that \method outperforms the state-of-the-art baselines on both classification and regression tasks.
\end{itemize}

\section{Related Work}

\subsection{Time series analysis}
Traditional methods for time series classification or regression include distance-based~\cite{Orsenigo2010CombiningDS,Grecki2015MultivariateTS}, 
feature-based~\cite{Schfer2015ScalableTS}, and ensemble methods~\cite{Lines2014TimeSC}. 
The distance-based methods measure the distance and similarity of (segments of) raw series data with some metric, 
like the Euclidean distance or \dtw (DTW)~\cite{keogh2005xacet}.
The feature-based methods capture the global/local patterns in the time series rely on a set of manually extracted or learned features.
The ensemble methods combine multiple weak classifiers to boost the performance of the models. 

However, involving heavy crafting on data pre-processing and labor-intensive feature engineering makes those methods struggling to cope with large-scale data and the performance is also limited for more complex patterns. 
Recurrent Neural Networks (RNNs), such as Gated Recurrent Unit~(GRU) and Long Short Term Memory~(LSTM)
have been popular due to their ability to extract high quality features automatically and handle the long term dependencies.
These deep methods tackle TS classification or forecasting by leveraging attention mechanisms~\cite{lai2018modeling,qin2017dual} or 
tensor factorization~\cite{sen2019think} for capturing shared information among series. Another trend is to model uncertainty 
by combining deep learning and State space models~\cite{salinas2020deepar}. 
Furthermore, some methods~\cite{salinas2020deepar,vincent2019shape, leguen20stripe} adopt seq2seq models for multi-step predictions.
Unlike those methods based on statistical views, 
our \method models TS from the perspective of the distributions.
Recently, the Transformers structure~\cite{vaswani2017attention} is proposed for sequence learning
with the self-attention mechanism. 
The vanilla Transformer can be easily modified for time series prediction, while it still fails for distribution matching.

Time series segmentation and clustering are two similar topics to our \tsqfull.
Time series segmentation~\cite{liu2008novel,chung2004evolutionary} aim to separate time series into several pieces that can be used to discover the underlying patterns; 
it mainly uses change-point detection algorithms, include sliding windows, bottom-up, and top-down methods.
The methods in segmentation almost don't utilize the distribution matching scheme and can't adapt to our problem. 
Time-series clustering~\cite{hallac2017toeplitz,ristanoski2013time}, on the other hand, aims at finding different groups consisting of similar (segments of) time series, 
which is clearly different from our situation.

\subsection{Distribution matching}
When the training and test data are coming from different distributions, it is common practice to adapt some domain adaptation (DA) algorithms to bridge their distribution gap, such that domain-invariant representations can be learned.
DA often performs instance re-weighting or feature transfer to reduce the distribution divergence in training and test data~\cite{wang2020transfer,zhu2020deep,wang2018visual,Ganin2016DomainAdversarialTO,Tzeng2017AdversarialDD}.
Similar to DA in general methodology, domain generalization (DG)~\cite{wang2021generalizing} also learns a domain-invariant model on multiple source domains in the hope that it can generalize well to the target domain~\cite{balaji2018metareg,li2018domain,muandet2013domain}.
The only difference between DA and DG is that DA assumes the test data is accessible while DG does not.
It is clear that our problem settings are different from DA and DG.
Firstly, DA and DG does not built for temporal distribution characterization since the domains are given a prior in their problems.
Secondly, most DA and DG methods are for classification tasks using CNNs than RNNs.
Thus, they might fail to capture the long-range temporal dependency in time series.
Several methods applied pre-training to RNNs~\cite{Cui2019TransferLF,chen2019transfer,Gupta2018TransferLF,Fawaz2018TransferLF,Yang2017TransferLF} which is similar to CNN-based pre-training.


\begin{figure*}[t!]
	\centering
	\vspace{-.1in}
	\subfigure[Overview of \method]{
		\centering
		\includegraphics[width=.16\textwidth]{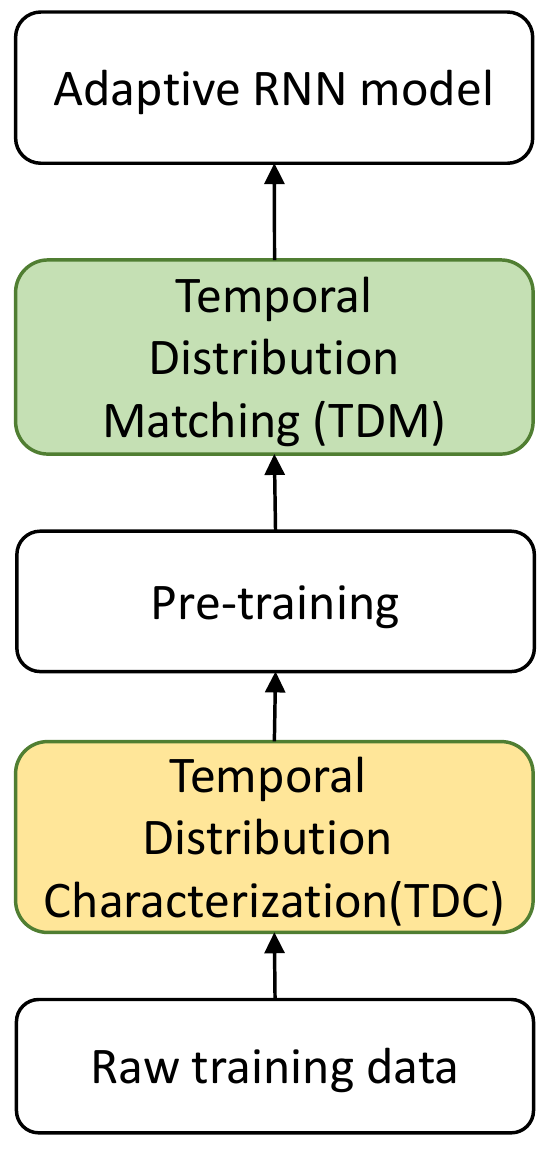}
		\label{fig-framework-all}}
	\subfigure[Temporal distribution characterization (\ttq)]{
		\centering
		\includegraphics[width=.2\textwidth]{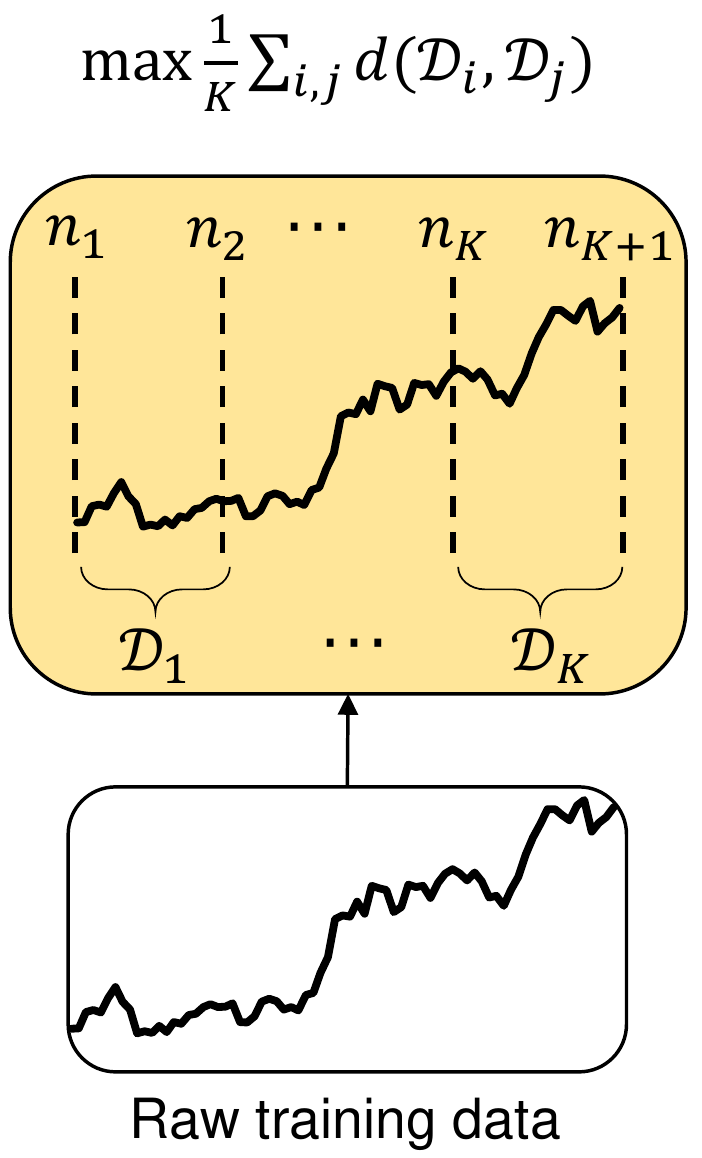}
		\label{fig-framework-tsq}}
	\subfigure[Temporal distribution matching (\cea)]{
		\centering
		\includegraphics[width=.47\textwidth]{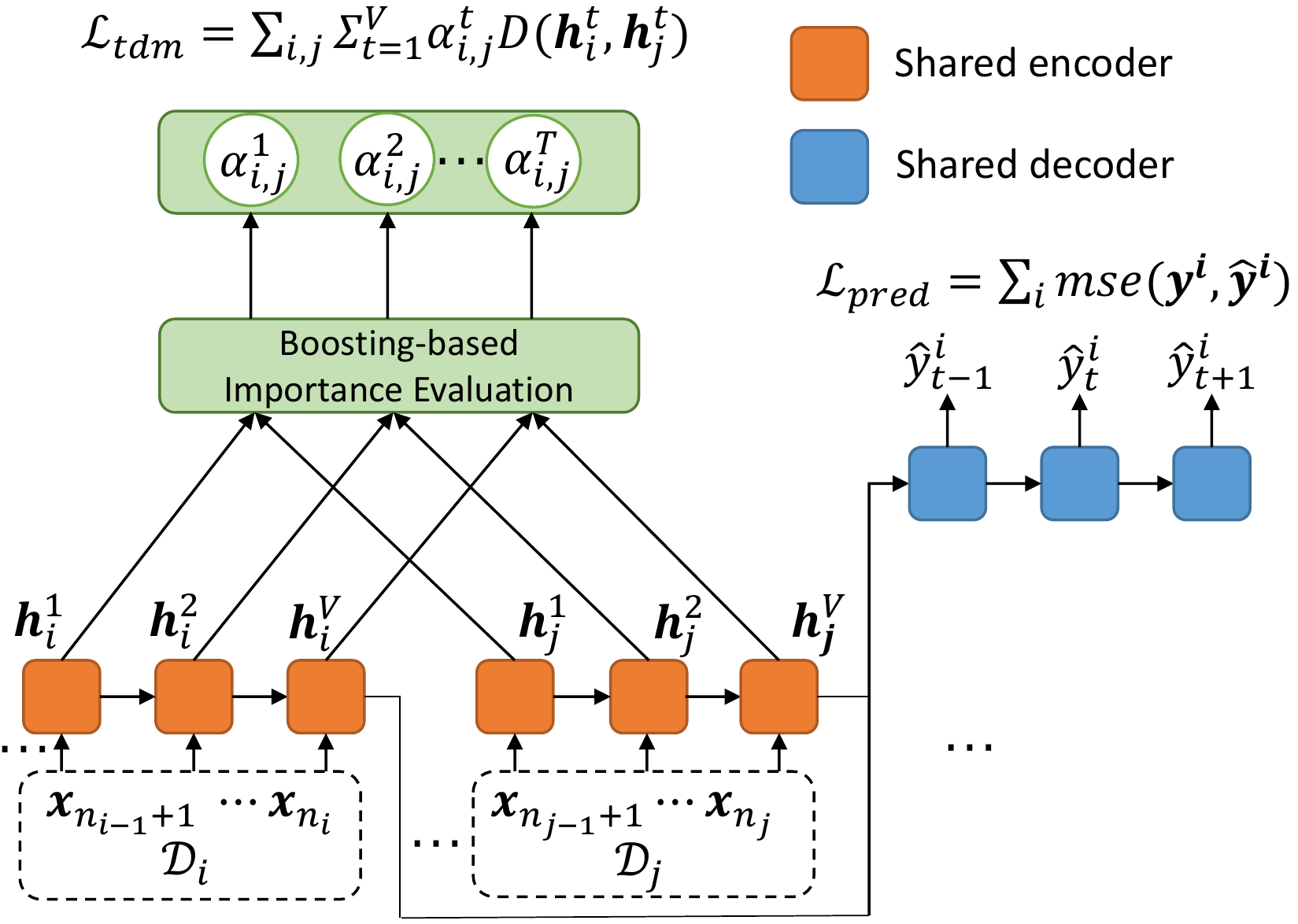}
		\label{fig-framework-tdm}}
	\vspace{-.15in}
	\caption{The framework of \method.}
	\label{fig-analysis}
	\vspace{-.1in}
\end{figure*}

\section{Problem Formulation} 
\label{sec-definition}


\begin{problem}[$r$-step ahead prediction] 
   \label{pro:r-step_pred}
   
	\textbf{Given} a time series of $n$ segments 
	$\mathcal{D}=\{\mathbf{x}_i, \mathbf{y}_i \}_{i=1}^{n}$, where $\mathbf{x}_i = \{x_i^1, \cdots, x_i^{m_i}\} \in \mathbb{R}^{p \times m_i}$ is a $p$-variate segment of length $m_i$\footnote{In general, $m_i$ may be different for different time segments.}, 
	and $\mathbf{y}_i = (y_i^1, \ldots, y_i^c) \in \mathbb{R}^{c}$ is the corresponding label. 
     
     \textbf{Learn} a precise prediction model $\mathcal{M}: \mathbf{x}_i \rightarrow \mathbf{y}_i$ 
     on the future $r \in \mathbb{N}^+$ steps \footnote{When $r=1$, it becomes the one-step prediction problem.} for segments $\{\mathbf{x}_j\}_{j=n+1}^{n+r}$ in the same time series. 
\end{problem}

In most existing prediction approaches, 
all time series segments, $\{\mathbf{x}_i\}_{i=1}^{n+r},$ 
are assumed to follow the same data distribution; 
a prediction model learned from all labeled training segments are supposed to perform well on the unlabeled testing segments. However, this assumption is too strong to hold in many real-world applications. In this work, we consider a more practical problem setting: the distributions of the training segments and the test segments can be different, and the distributions of the training segments can also be changed over time as shown in \figurename~\ref{fig-motivation}. In the following, we formally define our problem. 

\begin{definition}[Covariate Shift~\cite{shimodaira2000improving}] 
	Given the training and the test data from two distributions $P_{train}(\mathbf{x},y), P_{test}(\mathbf{x},y)$, covariate shift is referred to the case that the marginal probability distributions are different, and the conditional distributions are the same, i.e., $P_{train}(\mathbf{x}) \ne P_{test}(\mathbf{x})$, and  $P_{train}(y|\mathbf{x}) = P_{test}(y|\mathbf{x})$.
\end{definition}

Note that the definition of covariate shift is for non-time series data. We now extend its definition for time series data as follows, which is referred to as \emph{temporal covariate shift} (TCS) in this work.



\begin{definition}[Temporal Covariate Shift] 
	Given a time series data $\mathcal{D}$ with $n$ labeled segments. Suppose it can be split into $K$ periods or intervals, i.e., $\mathcal{D}=\{\mathcal{D}_1, \cdots, \mathcal{D}_K\}$, 
	where $\mathcal{D}_k = \{\mathbf{x}_i, \mathbf{y}_i\}^{n_{k+1}}_{i=n_k+1}$, $n_1 = 0$ and $n_{k+1} = n$.
	Temporal Covariate Shift (TCS) is referred to the case that all the segments in the same period $i$ follow the same data distribution $P_{ \mathcal{D}_i}(\mathbf{x},y)$, while for different time periods $1 \le i \ne j \le K$, $P_{ \mathcal{D}_i}(\mathbf{x}) \ne P_{ \mathcal{D}_j}(\mathbf{x})$ and 
	$P_{ \mathcal{D}_i}(y|\mathbf{x}) = P_{ \mathcal{D}_j}(y|\mathbf{x})$.
\end{definition}


To learn a prediction model with good generalization performance under TCS, a crucial research issue is to capture the \emph{common knowledge} shared among different periods of $\mathcal{D}$~\cite{kuznetsov2014generalization}. Take the stock price prediction as an example. Though the financial factors ($P(x)$) vary with the change of market (i.e., different distributions in different periods of time), some common knowledge, e.g., the economic laws and patterns ($P(y|x)$) in different periods, can still be utilized to make precise predictions over time. 


However, the number of periods $K$ and the boundaries of each period under TCS are usually unknown in practice. Therefore, before designing a model to capture commonality among different periods in training, we need to first discover the periods by comparing their underlying data distributions such that segments in the same period follow the same data distributions. The formal definition of the problem studied in this work is described as follows.

\begin{problem}[$r$-step TS prediction under TCS]
	Given a time series of $n$ segments with corresponding labels, $\mathcal{D}=\{\mathbf{x}_i, \mathbf{y}_i \}_{i=1}^{n}$, for training. Suppose there exist an unknown number of $K$ underlying periods in the time series, so that segments in the same period $i$ follow the same data distribution $P_{ \mathcal{D}_i}(\mathbf{x},y)$, while for different time periods $1 \le i \ne j \le K$, $P_{ \mathcal{D}_i}(\mathbf{x}) \ne P_{ \mathcal{D}_j}(\mathbf{x})$ and 
	$P_{ \mathcal{D}_i}(y|\mathbf{x}) = P_{ \mathcal{D}_j}(y|\mathbf{x})$. Our goal is to automatically discover the $K$ periods in the training time series data, and learn a prediction model $\mathcal{M}: \mathbf{x}_i \rightarrow \mathbf{y}_i$ by exploiting the commonality among different time periods, such that it makes precise predictions on the future $r$ segments, $\mathcal{D}_{tst} = \{\mathbf{x}_j\}_{j=n+1}^{n+r}$. Here we assume that the test segments are in the same time period, $P_{ \mathcal{D}_{tst}}(\mathbf{x}) \ne P_{ \mathcal{D}_i}(\mathbf{x})$ and
	$P_{ \mathcal{D}_{tst}}(y|\mathbf{x}) = P_{ \mathcal{D}_i}(y|\mathbf{x})$ for any $1 \le i \le K$. 
\end{problem}

Note that learning the periods underlying the training time series data is challenging as both the number of periods and the corresponding boundaries are unknown, and the search space is huge. Afterwards, with the identified periods, the next challenge is how to learn a good generalizable prediction model \predmodel to be used in the future periods. If the periods are not discovered accurately, it may affect the generalization performance of the final prediction model.

\section{Our proposed model: \method}

In this section, we present our proposed framework, termed Adaptive RNNs (\method), to solve the temporal covariate shift problem.
\figurename~\ref{fig-framework-all} depicts the overview of \method. 
As we can see, it mainly consists of two novel algorithms: 
\begin{compactitem}
	\item \emph{Temporal Distribution Characterization (\ttq):} to characterize the distribution information in TS; 
	\item \emph{Temporal Distribution Matching (\cea):} to match the distributions of the discovered periods to build a time series prediction model \predmodel.
\end{compactitem}

Given the training data, 
\method first utilizes \ttq to split it into the periods which fully characterize its distribution information. It then applies the \cea module to perform distribution matching among the periods to build a generalized prediction model \predmodel.
The learned \predmodel is finally used to make $r$-step predictions on new data.

The rationale behind \method is as follows. In \ttq, the model \predmodel is expected to work under the \emph{worst} distribution scenario where the distribution gaps among the different periods are large, thus the optimal split of periods can be determined by \emph{maximizing their dissimilarity}.
In \cea, \predmodel utilizes the common knowledge of the learned time periods by matching their distributions via RNNs with a regularization term to make precise future predictions.

\subsection{Temporal distribution characterization}

Based on \emph{the principle of maximum entropy}~\cite{jaynes1982rationale}, maximizing the utilization of shared knowledge underlying a times series under temporal covariate shift can be done by finding periods which are most dissimilar to each other, which is also considered as the worst case of temporal covariate shift since the cross-period distributions are the most diverse. 
Therefore, as \figurename~\ref{fig-framework-tsq} shows, \ttq achieves this goal for splitting the TS by solving an optimization problem whose objective can be formulated as:

\begin{equation}
	\label{equ:ttq_obj}
	\begin{split}
		\max_{0 < K \le K_0} \, {\max_{n_1, \cdots, n_K}   \frac{1}{K}\sum_{1 \le i \ne j \le K} d(\mathcal{D}_i, \mathcal{D}_j)}\\
		\mathrm{s.t.} \,\, \forall  \, i, \, \Delta_1 < |\mathcal{D}_i| < \Delta_2; \, \sum_{i} |\mathcal{D}_i| = n
	\end{split}
\end{equation}
where $d$ is a distance metric, 
$\Delta_1$ and $\Delta_2$ are predefined parameters to avoid trivial solutions (e.g., very small values or very large values may fail to capture the distribution information), 
and $K_0$ is the hyperparameter to avoid over-splitting. 
The metric $d(\cdot,\cdot)$ in Eq.~\eqref{equ:ttq_obj} can be any distance function, e.g., Euclidean or Editing distance, or some distribution-based distance / divergence, like MMD~\cite{Gretton2012OptimalKC} and KL-divergence.
We will introduce how to select $d(\cdot, \cdot)$ in Section~\ref{sec-method-dist}.

The learning goal of the optimization problem~\eqref{equ:ttq_obj} is to maximize the averaged period-wise distribution distances by searching $K$ and the corresponding periods so that the distributions of each period are as diverse as possible and the learned prediction model has better a more generalization ability.

We explain in detail for the splitting scheme in \eqref{equ:ttq_obj} with the principle of maximum entropy (ME).
Why we need the most dissimilar periods instead of the most similar ones?
According to ME, when no prior assumptions are made on the splitting of the time series data, it is reasonable that the distributions of each period become as diverse as possible to maximize the entropy of the total distributions.
This allows more general and flexible modeling of the future data.
Therefore, since we have no prior information on the test data which is unseen during training, it is more reasonable to train a model at the worst case, which can be simulated using diverse period distributions. If the model is able to learn from the worst case, it would have better generalization ability on unseen test data. This assumption has also been validated in theoretical analysis of time series models~\cite{kuznetsov2014generalization,oreshkin2020meta} that diversity matters in TS modeling.

In general, the time series splitting optimization problem in~\eqref{equ:ttq_obj} is computationally intractable and may not have a closed-form solution.
By selecting a proper distance metric, however, the optimization problem~\eqref{equ:ttq_obj} can be solved with dynamic programming (DP)~\cite{ross2014introduction}.
In this work, by considering the scalability and the efficiency issues on large-scale data, 
we resort to a greedy algorithm to solve~\eqref{equ:ttq_obj}.
Specifically, for efficient computation and avoiding trivial solutions, we evenly split the time series into $N=10$ parts, where each part is the minimal-unit period that cannot be split anymore. We then randomly search the value of $K$ in $\{2, 3, 5, 7, 10\}$.
Given $K$, we choose each period of length $n_j$ based on a greedy strategy.
Denote the start and the end points of the time series by $A$ and $B$, respectively.
We first consider $K=2$ by choosing $1$ splitting point (denote it as $C$) from the $9$ candidate splitting points via maximizing the distribution distance $d(S_{AC}, S_{CB})$.
After $C$ is determined, we consider $K=3$ and use the same strategy to select another point $D$. A similar strategy is applied to different values of $K$.   
Our experiments show that this algorithm can select the more appropriate periods than random splits. 
Also, it shows that when $K$ is very large or very small, the final performance of the prediction model becomes poor (Section~\ref{sec-exp-tdc}).

\subsection{Temporal distribution matching}
Given the learned time periods, the TDM module is designed to learn the common knowledge shared by different periods via matching their distributions.
Thus, the learned model $\mathcal{M}$ is expected to generalize well on unseen test data compared with the methods which only rely on local or statistical information.

The loss of TDM for prediction, $\mathcal{L}_{pred}$, can be formulated as:
\begin{equation}
\label{eq-mse}
    \mathcal{L}_{pred}(\theta) = \frac{1}{K}\sum_{j=1}^{K} \frac{1}{|\mathcal{D}_j|}\sum_{i=1}^{|\mathcal{D}_j|} \ell (\mathbf{y}_i^{j}, \mathcal{M}(\mathbf{x}_i^{j};\theta)),
\end{equation}
where $(\mathbf{x}_i^{j}, \mathbf{y}^{j}_i)$ denotes the $i$-th labeled segment from period $\mathcal{D}_j$, $\ell(\cdot, \cdot)$ is a loss function, e.g., the MSE loss, and $\theta$ denotes the learnable model parameters.

However, minimizing~\eqref{eq-mse} can only learn the predictive knowledge from each period, which cannot reduce the distribution diversities among different periods to harness the common knowledge.
A na\"ive solution is to match the distributions in each period-pair $\mathcal{D}_i$ and $\mathcal{D}_j$ by adopting some popular distribution matching distance ($d(\cdot, \cdot)$) as the regularizer.
Following existing work on domain adaptation~\cite{Ganin2016DomainAdversarialTO,Sun2016DeepCC} that distribution matching is usually performed on high-level representations, we apply the distribution matching regularization term on the final outputs of the cell of RNN models.
Formally, we use $\mathbf{H} = \{\mathbf{h}^t\}^{V}_{t=1} \in \mathbb{R}^{V \times q}$ to denote the $V$ hidden states of an RNN with feature dimension $q$.
Then, the period-wise distribution matching on the final hidden states for 
a pair $(\mathcal{D}_i, \mathcal{D}_j)$ can be represented as:
\begin{equation*}
\label{eq-naivedist}
    \mathcal{L}_{dm}(\mathcal{D}_i, \mathcal{D}_j; \theta) =  d(\mathbf{h}_i^{V}, \mathbf{h}_j^{V};\theta).
\end{equation*}
Unfortunately, the above regularization term fails to capture the \emph{temporal dependency} of each hidden state in the RNN. 
As each hidden state only contains \emph{partial} distribution information of an input sequence, \emph{each} hidden state of the RNN should be considered while constructing a distribution matching regularizer as described in the following section.


\subsubsection{Temporal distribution matching}
As shown in \figurename~\ref{fig-framework-tdm}, we propose the \emph{Temporal Distribution Matching (TDM)} module to adaptively match the distributions 
between the RNN cells of two periods while capturing the temporal dependencies.
TDM introduces the \emph{importance vector} $\bm{\alpha} \in \mathbb{R}^{V}$ to learn the relative 
importance of $V$ hidden states inside the RNN, where all the hidden states are weighted with a normalized $\bm{\alpha}$.
Note that for each pair of periods, there is an $\bm{\alpha}$, and we omit the subscript if there is no confusion.
In this way, we can \emph{dynamically} reduce the distribution divergence of cross-periods.

Given a period-pair $(\mathcal{D}_i, \mathcal{D}_j)$, the loss of temporal distribution matching is formulated as:
\begin{equation}
	\label{eq-tdm}
	\mathcal{L}_{tdm}(\mathcal{D}_i, \mathcal{D}_j; \theta) = \sum_{t=1}^{V} \alpha_{i,j}^{t} d(\mathbf{h}^{t}_i , \mathbf{h}^{t}_j; \theta),
\end{equation}
where $\alpha_{i,j}^{t}$ denotes the distribution importance between the periods $\mathcal{D}_i$ and $\mathcal{D}_j$ at state $t$.

All the hidden states of the RNN can be easily computed by following the standard RNN computation. 
Denote by $\delta(\cdot)$ the computation of a next hidden state based on a previous state. The state computation can be formulated as
\begin{equation}
	\mathbf{h}_i^{t}=\delta(\mathbf{x}_i^t, \mathbf{h}^{t-1}_i).
\end{equation}

By integrating \eqref{eq-mse} and \eqref{eq-tdm}, the final objective of temporal distribution matching (one RNN layer) is:
\begin{equation}
\label{eq-all}
    \mathcal{L}(\theta, \bm{\alpha}) = \mathcal{L}_{pred}(\theta) +\lambda \frac{2}{K(K-1)} \sum_{i,j}^{i \ne j} \mathcal{L}_{tdm}(\mathcal{D}_i, \mathcal{D}_j; \theta, \bm{\alpha}),
\end{equation}
where $\lambda$ is a trade-off hyper-parameter.
Note that in the second term, we compute the average of the distribution distances of all pairwise periods.
For computation, we take a mini-batch of $\mathcal{D}_i$ and $\mathcal{D}_j$ to perform forward operation in RNN layers and concatenate all hidden features. Then, we can perform TDM using \eqref{eq-all}.

\subsubsection{Boosting-based importance evaluation}
We propose a \emph{Boosting-based importance evaluation} algorithm to learn $\bm{\alpha}$.
Before that, we first perform pre-training on the network parameter $\mathbf{\theta}$ using the fully-labeled data across all periods, i.e., use \eqref{eq-mse}. This is to learn better hidden state representations to facilitate the learning of $\bm{\alpha}$. We denote the pre-trained parameter as $\mathbf{\theta}_0$.

With $\mathbf{\theta}_0$, TDM uses a boosting-based~\cite{schapire2003boosting} procedure to learn the hidden state importance.
Initially, for each RNN layer, all the weights are initialized as the same value in the same layer, i.e., $\bm{\alpha}_{i,j} = \{1/V\}^{V}$. 
We choose the cross-domain distribution distance as the indicator of boosting. If the distribution distance of epoch $n+1$ is larger than that of epoch $n$, we increase the value of $\alpha_{i,j}^{t,(n+1)}$ to enlarge its effect for reducing the distribution divergence. Otherwise, we keep it unchanged. The boosting is represented as:
\begin{equation}
	\label{eq-local-alpha}
	\alpha_{i,j}^{t,(n+1)} =\left\{
	\begin{aligned}
		& \alpha_{i,j}^{t,(n)} \times G(d^{t,(n)}_{i,j}, d^{t,(n-1)}_{i,j}) &  & d^{t,(n)}_{i,j} \ge d^{t,(n-1)}_{i,j} \\
		& \alpha_{i,j}^{t,(n)}                                                                &  & \text{otherwise}                                       \\
	\end{aligned}
	\right.
\end{equation}
where
\begin{equation}
	G(d^{t,(n)}_{i,j}, d^{t,(n-1)}_{i,j}) = (1+\sigma(d^{t,(n)}_{i,j} - d^{t,(n-1)}_{i,j}))
\end{equation}
is the updating function computed by distribution matching loss at different learning stages. And $d^{t,(n)}_{i,j} = D(\mathbf{h}_i^{t},\mathbf{h}_j^{t}; \alpha_{i,j}^{t,(n)})$ is the distribution distance at time step $t$ in epoch $n$. $\sigma(\cdot)$ is the sigmoid function. It is easy to see that $G(\cdot, \cdot) > 1$, ensuring the increment of the importance.
Similar to naive method, we normalize it as $ \alpha_{i,j}^{t,(n+1)} = \frac{\alpha_{i,j}^{t,(n+1)}}{\sum_{j=1}^{V}\alpha_{i,j}^{t,(n+1)}}$.
By \eqref{eq-tdm} and \eqref{eq-local-alpha}, we can learn the value of $\bm{\alpha}$. 

\begin{remark}
The objective~\eqref{eq-tdm} is defined on one RNN layer, it can be easily implemented for multiple layers to achieve multi-layer distribution matching.
As for inference, we simply use the optimal network parameter $\mathbf{\theta}^\ast$. The computation of domain-wise distribution distances can also be made efficient following existing work~\cite{li2018domain}.
\end{remark}

\begin{remark}
\label{remark}
In fact, there is also a na\"ive approach to acquire $\bm{\alpha}$ by using another network with weight $\mathbf{W}_{i,j} \in \mathbb{R}^{2q \times V}$ that takes $(\mathbf{H}_i,\mathbf{H}_j)$ as inputs and output $\bm{\alpha}$. For instance, $\bm{\alpha}$ can be learned as $\bm{\alpha}_{i,j} = g(\mathbf{W}_{i,j} \odot [\mathbf{H}_{i}, \mathbf{H}_{j}]^\top$) where $g$ is an activation and normalization function and $\odot$ is element-wise production. After computing $\alpha_{i,j}$, we normalize it by a \textsf{softmax} function.
Unfortunately, this na\"ive approach does not work in reality for two reasons.
First, since $\bm{\alpha}_{i,j}$ and $\mathbf{\theta}$ are highly correlated and at early training stage, the hidden state representations 
learned by $\mathbf{\theta}$ are less meaningful, which will result in insufficient learning of $\bm{\alpha}_{i,j}$. 
Second, it is expensive to learn such $\mathbf{W}_{i,j}$ for each domain pair in RNN models. 
\end{remark}

\begin{algorithm}[t]
	\caption{\method}
	\label{algo-adarnn}
	\begin{algorithmic}[1]
		\REQUIRE A time series dataset $\mathcal{D}$, $\lambda$, max iteration number $Iter$
		\STATE Perform temporal distribution characterization by solving~\eqref{equ:ttq_obj} to get $K$ periods $\{\mathcal{D}_i\}$'s.
		\STATE End-to-end pre-training by minimizing~\eqref{eq-mse} with $\lambda=0$
		\STATE \textbf{For} iteration $i = 1 \text{ to } Iter$ :  
		\STATE \quad Learn the value of $\bm{\alpha}$ and $\theta$ by solving~\eqref{eq-all} and using~\eqref{eq-local-alpha}
		\STATE \textbf{End For}
		\RETURN Optimal parameters $\mathbf{\theta}^\ast, \bm{\alpha}^\ast$.
	\end{algorithmic}
\end{algorithm}

\subsection{Computation of the distribution distance}
\label{sec-method-dist}

\method is agnostic to RNN structures (vanilla RNNs, LSTMs, and GRUs) and the distribution distances, i.e., the function $d(\cdot, \cdot)$.
Technically, $d(\cdot, \cdot)$ in TDC and TDM can be the same choice. 
We adopt several popular distances: cosine distance, MMD~\cite{Gretton2012OptimalKC} and adversarial distance~\cite{Ganin2016DomainAdversarialTO}. 
Also note that the TDC and TDM are currently two separate stages for controlling computation complexity.
It shall be a future work to end-to-end optimize these two stages.
The complete process of \method is presented in Algorithm~\ref{algo-adarnn}.
The detailed implementations of these distances are presented in Appendix.

\section{Experiments}
\label{sec-exp}

\subsection{Setup}

We mainly use GRUs (Gated Recurrent Units)~\cite{chung2014empirical} as the base RNN cells for its high performance in real applications\footnote{It is natural to use simple RNNs or LSTMs.}. 
We test \method in four real-world datasets as summarized in \tablename~\ref{tb-dataset}.

\begin{table}[htbp]
	\centering
	\caption{Information of the datasets used in this paper.}
	\vspace{-.1in}
	\label{tb-dataset}
	\resizebox{.48\textwidth}{!}{
		\begin{tabular}{lrrrrc}	\toprule
			\textbf{Dataset} & \textbf{\#Train} & \textbf{\#Valid} & \textbf{\#Test} & \textbf{dim. ($p$)} & \textbf{Task} \\ \hline
			UCI activity & 6,000   & 1,352    & 2,947    & 128 & Classification \\ 
			Air quality & 29,232   & 2,904   & 2,832   & 6    & Regression \\
			Electric power & 1,235,964   & 413,280   & 413,202   & 7    & Regression \\
			Stock price & 778,200 & 271,690 & 504,870 & 360  & Regression \\ \bottomrule
		\end{tabular}
	}
	\vspace{-.1in}
\end{table}

We compare \method with four categories of methods: 
\begin{itemize}
	\item Traditional TS modeling methods, including ARIMA~\cite{zhang2003time}, FBProphet~\cite{fbprophet}, LightGBM, and GRU~\cite{chung2014empirical}; 
	\item Latest TS models, including LSTNet~\cite{lai2018modeling} and STRIPE~\cite{leguen20stripe}\footnote{STRIPE is an improved and stronger version of DILATE~\cite{vincent2019shape}.};
	\item Transformer~\cite{vaswani2017attention} by removing its mask operation;
	\item Variants of popular domain adaptation / generalization methods and we denote them as ``MMD-RNN'' and ``DANN-RNN''.
\end{itemize}

Note that the comparison is mainly for the final prediction performance using temporal distribution matching after obtaining the time periods using TDC, which the evaluation of TDC is in Section~\ref{sec-exp-tdc}.
Transformer~\cite{vaswani2017attention} was originally proposed for machine translation.
Therefore, we alter it by removing the masks for TS data by following implementation \footnote{\url{https://timeseriestransformer.readthedocs.io/en/latest/README.html}}.
We use $4$ encoder layers of Transformer with $8$ heads for self-attention tuned by cross-validation to ensure that it achieves the best performance on the datasets.
Existing DA / DG methods cannot be directly applied to our problem, since most of them are for CNN-based classification tasks.
Therefore, we use variants of them. Specifically, we use ``MMD-RNN'' to denote popular MMD-based methods such as MEDA~\cite{wang2018visual,wang2020transfer} and MMD-AAE~\cite{li2018domain}. We use ``DANN-RNN'' to denote popular domain-adversarial training methods such as RevGrad~\cite{Ganin2016DomainAdversarialTO} and DAAN~\cite{yu2019transfer}. These two distances are widely adopted in recent DA and DG literature.
We change their backbone network to be RNNs which are the same as other comparison methods and then add the MMD or adversarial distances to the final output of the RNN layers.

All methods use the same train-valid-test split and periods after TDC and the final results are averaged on five random runs.
Optimal hyperparameters are tuned on the validation set.
Note that to balance the efficiency and performance in \method, we limit the range of 
$K$ in $\{2,3,5,7,10\}$ to acquire the best splitting of periods. We set the pre-training epochs to be $10$ in TDM.
We use different distance function $d$ and the main results are based on the MMD~\cite{Gretton2012OptimalKC}.
Other distances are discussed later.
Code of \method is at \url{https://github.com/jindongwang/transferlearning/tree/master/code/deep/adarnn}.

\subsection{Human activity recognition}

The UCI smartphone activity~\cite{almaslukh2017effective} is a time-series dataset for classification task. 
It is collected by thirty volunteers wearing a smartphone to record the accelerometer, gyroscope, and magnetometer signals 
while performing six different activities (walking, upstairs, downstairs, sitting, standing, and lying). 
We use $7,352$ instances for training and $2,947$ instances as the test following~\cite{almaslukh2017effective}.
It is clear that training and testing sets have different distributions and TCS exists in the training set due to different activities and users.
We firstly clip the data using sliding window with the size of 128. 
As there are 9 channels for each instance, we regard it a sequence with length of 128 and feature of 9 dimension. 
For this task, we use accuracy (ACC), precision (P), recall (R), F1, and area under curve (AUC) as the evaluation metrics.

The network is composed of two-layer GRUs with the hidden state dimension $32$ and a FC layer with output to be $r=6$. We optimize by Adam optimizer with the learning rate of 0.005 with a batch size 64. We set $\lambda=0.1$.
On this dataset, we did not run the latest methods such as LSTNet and STRIPE as well as FBProphet and ARIMA since they are built for regression tasks. Transformer is not compared since we did not get good performance.

\tablename~\ref{tb-act} shows the results on this classification task.
It is obvious that \method achieves the best performance on all evaluation metrics that outperforms the adversarial-based DANN method by $\bm{2.56}\%$ in accuracy and $\bm{3.07}\%$ in F1 score.
This means that \method is effective in non-stationary time series classification tasks.

\begin{table}[t!]
	\centering
	\caption{Results on UCI time series classification dataset.}
	\label{tb-act}
	\vspace{-.1in}
	\resizebox{.4\textwidth}{!}{
		\begin{tabular}{lccccc}
			\toprule
			Method & ACC            & P              & R              & F1             & AUC            \\
			\hline
			LightGBM     & 84.11          & 83.73          & 83.63          & 84.91          & 90.23          \\
			GRU          & 85.68          & 85.62          & 85.51          & 85.46          & 91.33          \\
			MMD-RNN       & 86.39          & 86.80          & 86.26          & 86.38          & 91.77          \\
			DANN-RNN        & 85.88          & 85.59          & 85.62          & 85.56          & 91.41          \\
			\method & \textbf{88.44} & \textbf{88.71} & \textbf{88.59} & \textbf{88.63} & \textbf{93.19} \\
			\bottomrule
		\end{tabular}
	}
	\vspace{-.2in}
\end{table}

\begin{table*}[t!]
	\centering
	\caption{Results on air quality prediction (left) and household electric power consumption prediction (right).}
	\vspace{-.1in}
	\label{tb-air-quality}
	\resizebox{.7\textwidth}{!}{
		\begin{tabular}{lccccccccr||r}
			\toprule
			& \multicolumn{2}{c}{Dongsi} & \multicolumn{2}{c}{Tiantan} & \multicolumn{2}{c}{Nongzhanguan} & \multicolumn{2}{c}{Dingling} & \multirow{2}{*}{$\Delta (\%)$} & \multirow{2}{*}{Electric Power}\\ 
			& RMSE & MAE & RMSE & MAE & RMSE & MAE & RMSE & MAE &  \\ \midrule
			FBProphet~\cite{fbprophet} & 0.1866 & 0.1403 & 0.1434 & 0.1119 & 0.1551 & 0.1221 & 0.0932 & 0.0736 & - & 0.080\\ 
			ARIMA & 0.1811 & 0.1356 & 0.1414 & 0.1082 & 0.1557 & 0.1156 & 0.0922 & 0.0709 & - & -\\ 
			GRU & 0.0510 & 0.0380 & 0.0475 & 0.0348 & 0.0459 & 0.0330 & 0.0347 & 0.0244 & 0.00 & 0.093\\ 
			MMD-RNN & 0.0360 & 0.0267 & 0.0183 & 0.0133 & 0.0267 & 0.0197 & 0.0288 & 0.0168 & -61.31 & 0.082\\ 
			DANN-RNN & 0.0356 & 0.0255 & 0.0214 & 0.0157 & 0.0274 & 0.0203 & 0.0291 & 0.0211 & -59.97 & 0.080\\ 
			LightGBM & 0.0587 & 0.0390 & 0.0412 & 0.0289 & 0.0436 & 0.0319 & 0.0322 & 0.0210 & -11.08 & 0.080\\ 
			LSTNet~\cite{lai2018modeling} & 0.0544 & 0.0651 & 0.0519 & 0.0651 & 0.0548 & 0.0696 & 0.0599 & 0.0705 & - & 0.080\\
			Transformer~\cite{vaswani2017attention} & 0.0339 & 0.0220 & 0.0233 & 0.0164 &	0.0226 & 0.0181 & 0.0263 &	0.0163 & -61.20 & 0.079\\  
			STRIPE~\cite{leguen20stripe} & 0.0365 & 0.0216 & 0.0204 & 0.0148 & 0.0248 & 0.0154 & 0.0304 & 0.0139 & -64.60 & 0.086\\ 
			\method & \textbf{0.0295} & \textbf{0.0185} & \textbf{0.0164} & \textbf{0.0112} & \textbf{ 0.0196} & \textbf{0.0122} & \textbf{0.0233} & \textbf{0.0150} & \textbf{-73.57} & \textbf{0.077}\\ 
			\bottomrule
		\end{tabular}
	}
   \vspace{-.1in}
\end{table*}

\begin{figure*}[t!]
	\centering
	\subfigure[\#Domain in TDC]{
		\centering
		\includegraphics[width=.23\textwidth]{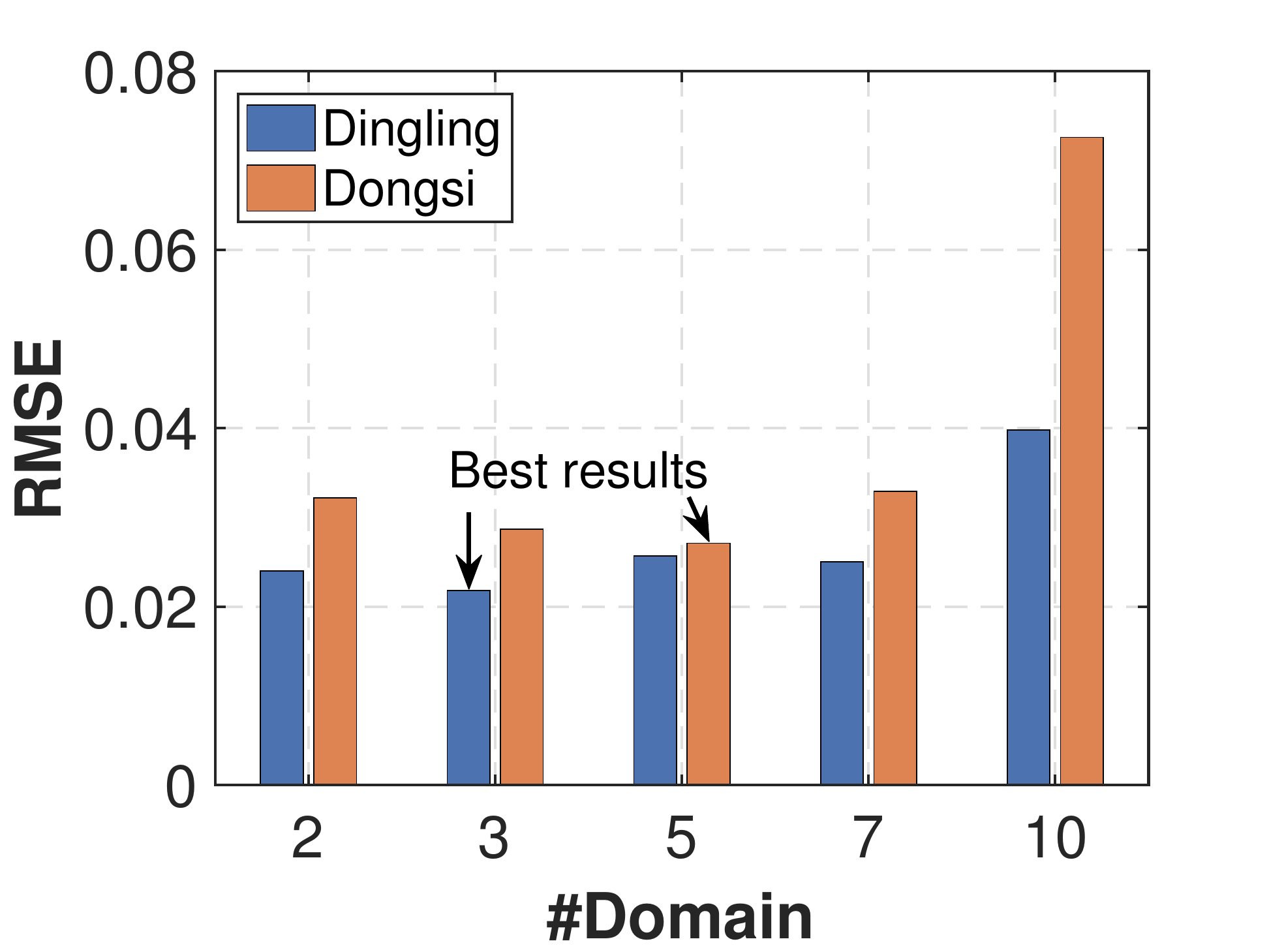}
		\label{fig-sub-domain}}
	\subfigure[Different domain split]{
		\centering
		\includegraphics[width=.23\textwidth]{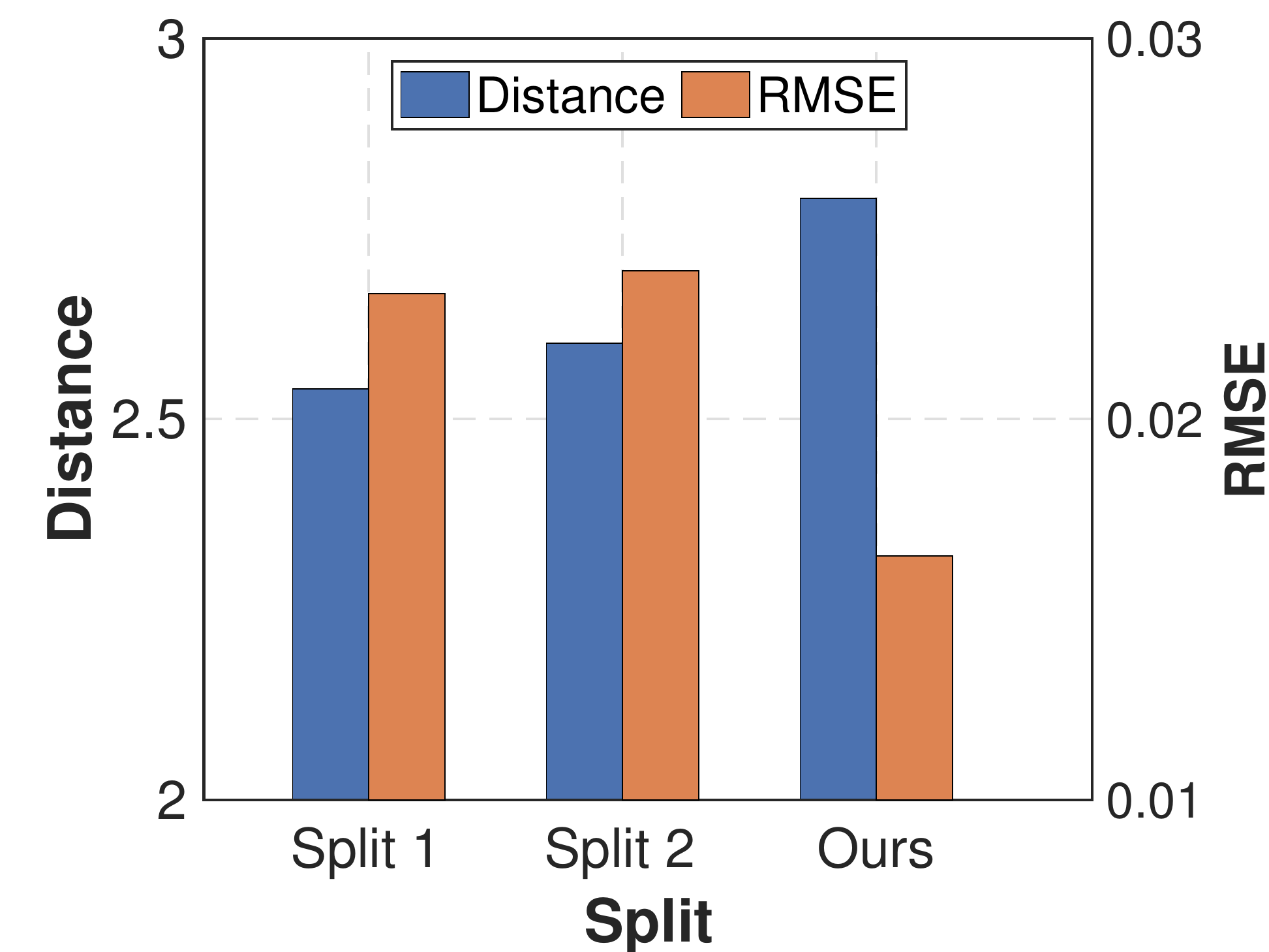}
		\label{fig-sub-tsq}}
	\subfigure[Temporal distribution matching]{
		\centering
		\includegraphics[width=.23\textwidth]{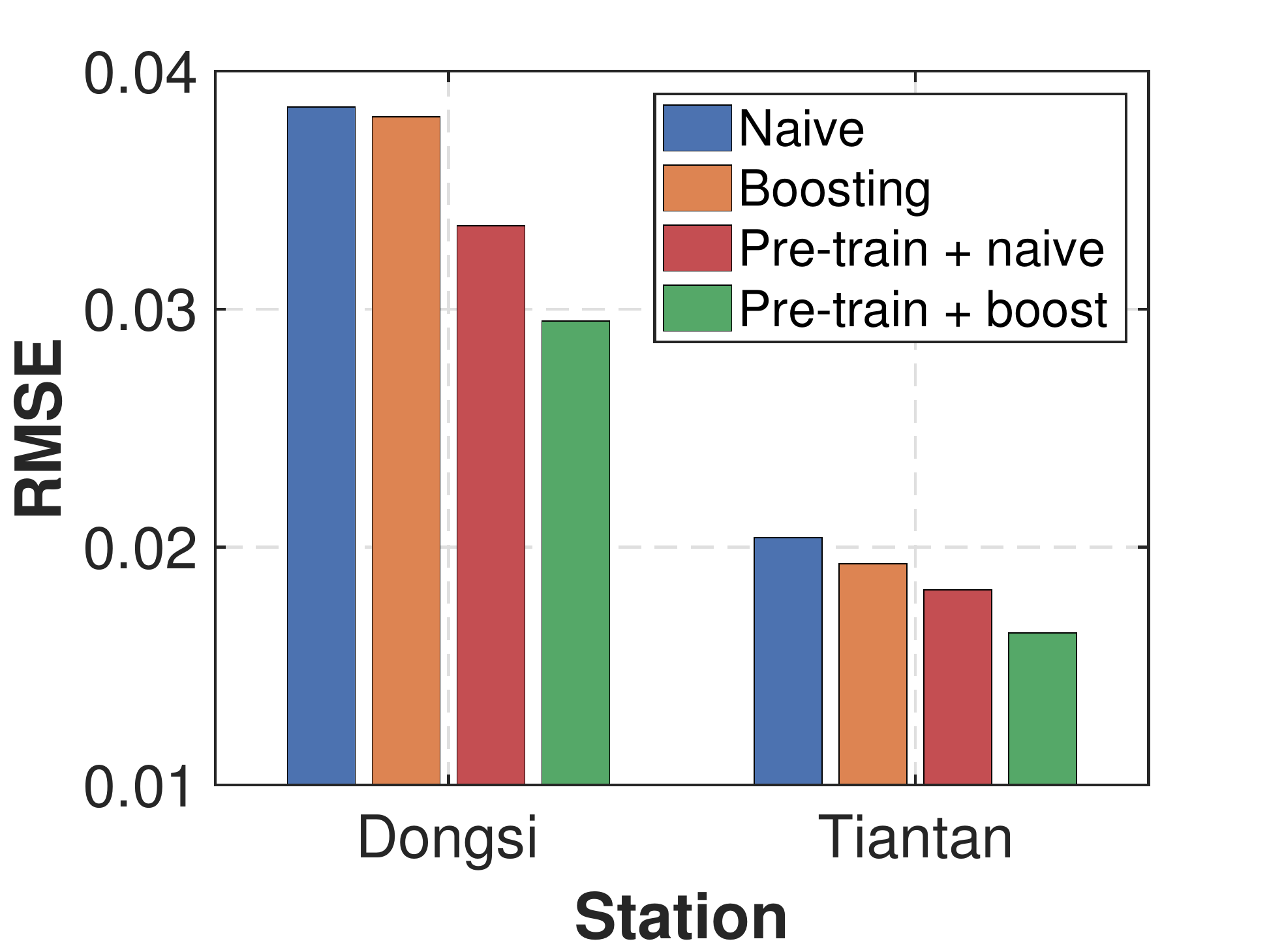}
		\label{fig-sub-tdm}}
	\subfigure[Effectiveness of $\alpha$]{
		\centering
		\includegraphics[width=.23\textwidth]{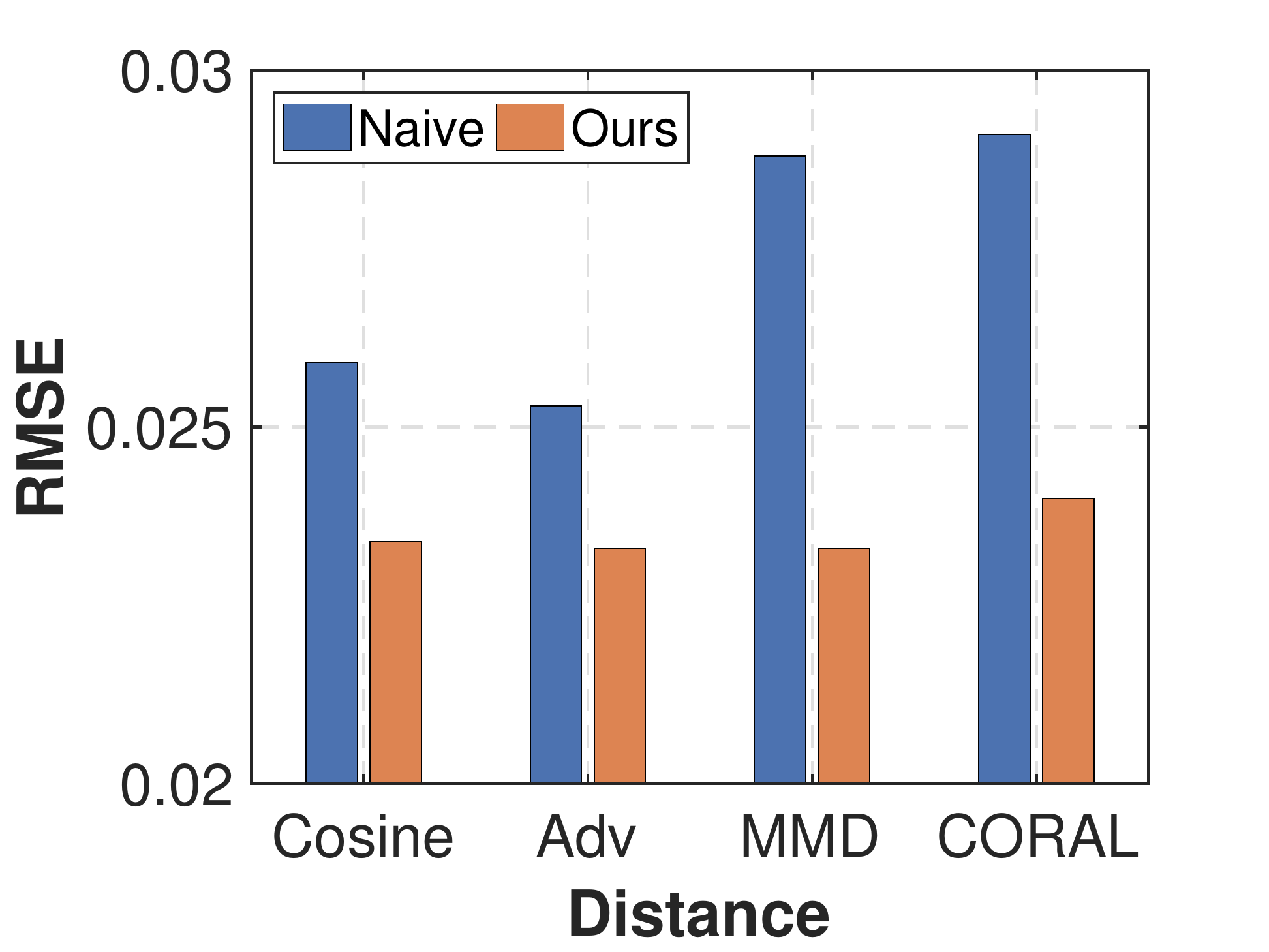}
		\label{fig-sub-alpha}}
	
	\vspace{-.1in}
	\caption{Ablation study. (a) Analysis of $K$ for \tsqfull. (b) Different domain split strategies. (c) Comparison of w. or w./o. adaptive weights in distribution matching. (d) Effectiveness of temporal distribution matching.}
	\label{fig-ablation}
	\vspace{-.1in}
\end{figure*}

\subsection{Air quality prediction}

The air-quality dataset~\cite{zhang2017cautionary} contains hourly air quality information 
collected from 12 stations in Beijing from 03/2013 to 02/2017. We randomly chose 
four stations (Dongsi, Tiantan, Nongzhanguan, and Dingling) and select six features (PM2.5, PM10, S02, NO2, CO, and O3). 
Since there are some missing data, we simply fill the empty slots using averaged values. 
Then, the dataset is normalized before feeding into the network to scale all features into the same range. 
This process is accomplished by max-min normalization and ranges data between 0 and 1.

The network is composed of two-layer GRUs with the dimension of the hidden state to be 64 and two bottleneck layers 
which contains two FC layers, and an FC layer with output to be $r = 1$. We set $\lambda=0.5$.
We optimize all the models by Adam optimizer with the learning rate of 0.005, we set the batch size to be 36.

\tablename~\ref{tb-air-quality} presents the results where the last column ``$\Delta (\%)$'' denotes the average increment over the baseline GRU. 
It can be seen that the proposed method achieves the best results in all four stations. 
Our \method significantly outperforms the strong 
baseline STRIPE~\cite{leguen20stripe} by $\bm{8.97}\%$ in RMSE reduction.

\subsection{Household electric power consumption}

This dataset~\cite{householdpower} contains 2,075,260 measurements gathered between 16/12/2006 and 26/11/2010 with a one-minute sampling rate.
It contains several attributes (global active power, global reactive power, voltage, global intensity, and three substrings). We get 2,062,346
measurements for the whole dataset after removing all the null data.
We construct the train/valid/test sets with a ratio of $6:2:2$, i.e., train on data from 16/12/2006 to 24/04/2009 and test on data from 25/02/2010 to 26/11/2010 and the rest is valid set.
We use the data of 5 consecutive days to predict the value of next day.

The network is composed of two-layer GRUs with the dimension set to be 64 and two bottleneck layers which contains two FC layers, and a FC layer with output to be $r=1$. We set $\lambda=0.001$.
We optimize all the models by Adam optimizer with the learning rate of 0.0005 and batch size of 36.

The right part of \tablename~\ref{tb-air-quality} shows the RMSE of predicting the household power consumption using different methods. Again, our \method achieves the best performance.

\subsection{Stock price prediction}

We further evaluate the performance of \method on a large private finance dataset where the task is to predict the stock price given the historical data from 01/2007 to 12/2019. This dataset contains over 2.8M samples with 360 financial factors as the features. We use the data from 01/2017 to 12/2019 as the test set, and the data from 01/2007 to 12/2014 as the training set with the rest as the validation set. We use information coefficient (IC) and information ratio (IR) as the evaluation metrics following existing work~\cite{kohara1997stock}. The detailed description and computation of these metrics are in appendix. 

The network is composed of two-layer GRUs with the dimension of hidden state to be 64 and 
two bottleneck layers which contains one FC layers, and a FC layer with output to be $r=1$. 
We optimize all the models by Adam optimizer with the learning rate of 0.0002, we set the batch size to be 800 and $\lambda=0.5$.

\tablename~\ref{tb-res-stock} shows that our proposed \method outperforms other comparison methods in all metrics.
While achieving the best IC values, its RankIC can still be higher than other methods. This means that \method is relatively steady in real applications.

\begin{table}[htbp]
	\centering
	\caption{Results on stock price prediction.}
	\vspace{-.1in}
	\label{tb-res-stock}
	\resizebox{.48\textwidth}{!}{
		\begin{tabular}{lcccccc}
			\toprule
			Method & IC & ICIR & RankIC & RankICIR & RawIC & RawICIR \\ \hline
			LightGBM & 0.063  & 0.551 & 0.056  & 0.502  & 0.064  & 0.568   \\
			GRU & 0.106 & 0.965 & 0.098 & 0.925 & 0.109 & 0.986 \\
			MMD-RNN & 0.107 & 0.962 & 0.101 & 0.926 & 0.108 & 0.967 \\
			DANN-RNN & 0.106 & 0.964 & 0.103 & 0.924 & 0.107 & 0.966 \\
			Transformer~\cite{vaswani2017attention} & 0.104 & 0.935 & 0.099 & 0.920 & 0.101 & 0.946 \\
			STRIPE~\cite{leguen20stripe} & 0.108 & 0.987  & 0.101   & 0.946    & 0.109 & 0.993  \\
			\method & \textbf{0.115} & \textbf{1.071} & \textbf{0.110} & \textbf{1.035} & \textbf{0.116} & \textbf{1.077} \\ \bottomrule
		\end{tabular}
	}
	\vspace{-.2in}
\end{table}

\subsection{Ablation study}

\subsubsection{Temporal distribution characterization}
\label{sec-exp-tdc}

To evaluate TDC, we design two experiments. Firstly, we compare the results of $K=2,3,5,7,10$ to analyze the importance of $K$ in Figure~\ref{fig-sub-domain}. We see that the performance is changing with the increment of $K$ and the best result is achieved when $K=3,5$ for two stations on air quality prediction, respectively. Then, when $K$ continues to increase, the performance drops. It means that $K$ is important in characterizing the distribution information in non-stationary TS.

Secondly, \figurename~\ref{fig-sub-tsq} shows the comparison of domain split algorithms using the same distribution matching method when $K=2$.
Here, ``Split 1'' denotes a random split and ``Split 2'' denotes the split where all periods are with similar distributions (i.e., just the opposite of TDC).
The ``distance'' denotes the distribution distance.
The results indicate that our TDC achieves the best performance in RMSE with the largest distribution distance.
The results of other values of $K$ are also the same.
Therefore, it is important that we split the periods to the \emph{worst} case when the distributions are the most diverse.
Both experiments show that TDC could effectively characterize the distribution information in time series.
Evaluation of TDC using other distances are in appendix.

\begin{figure*}[t!]
	\centering
	\vspace{-.1in}
	\hspace{-.2in}
	\subfigure[Different distribution distance]{
		\centering
		\includegraphics[width=.19\textwidth]{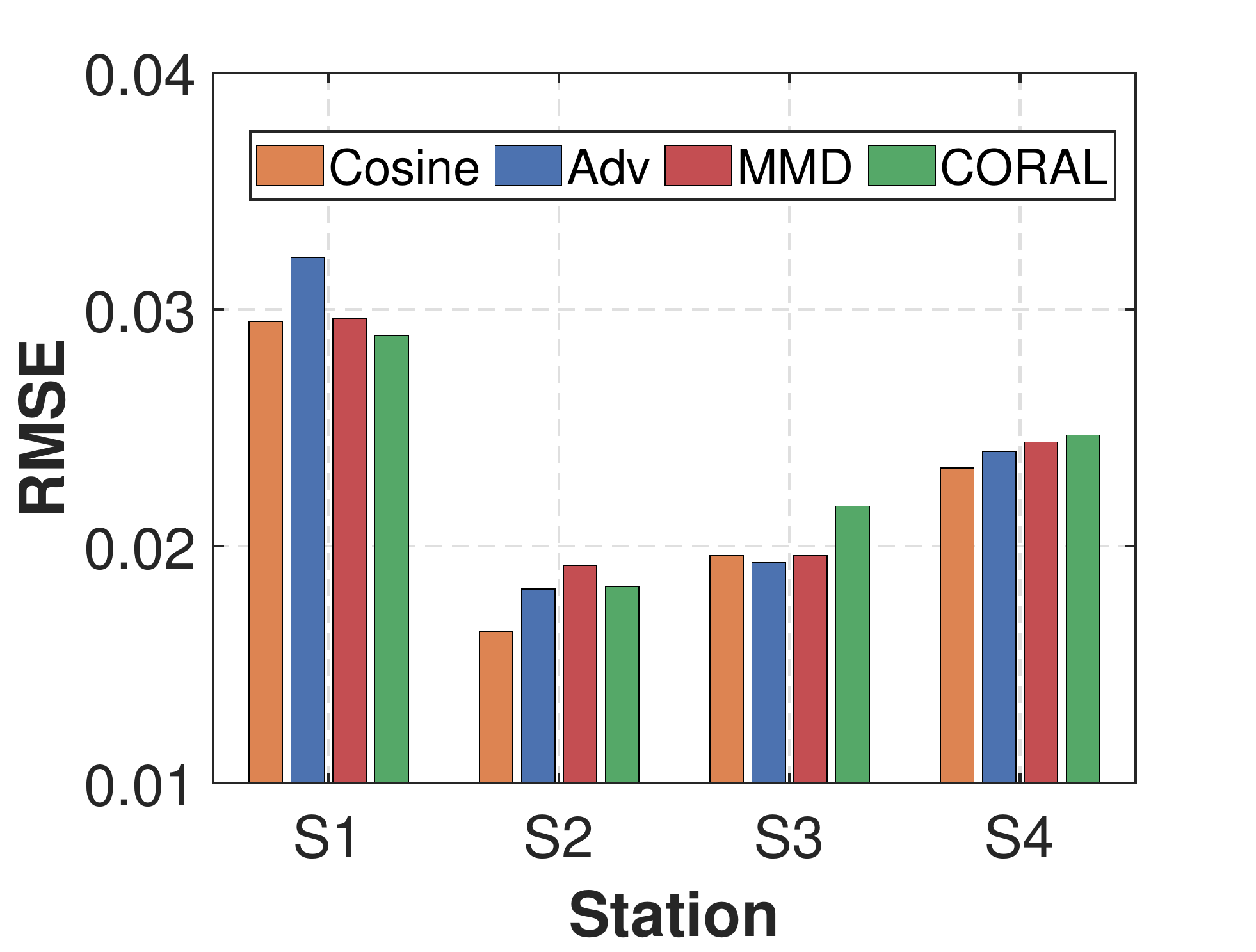}
		\label{fig-sub-distance}}
	\subfigure[Change of distance]{
		\centering
		\includegraphics[width=.19\linewidth]{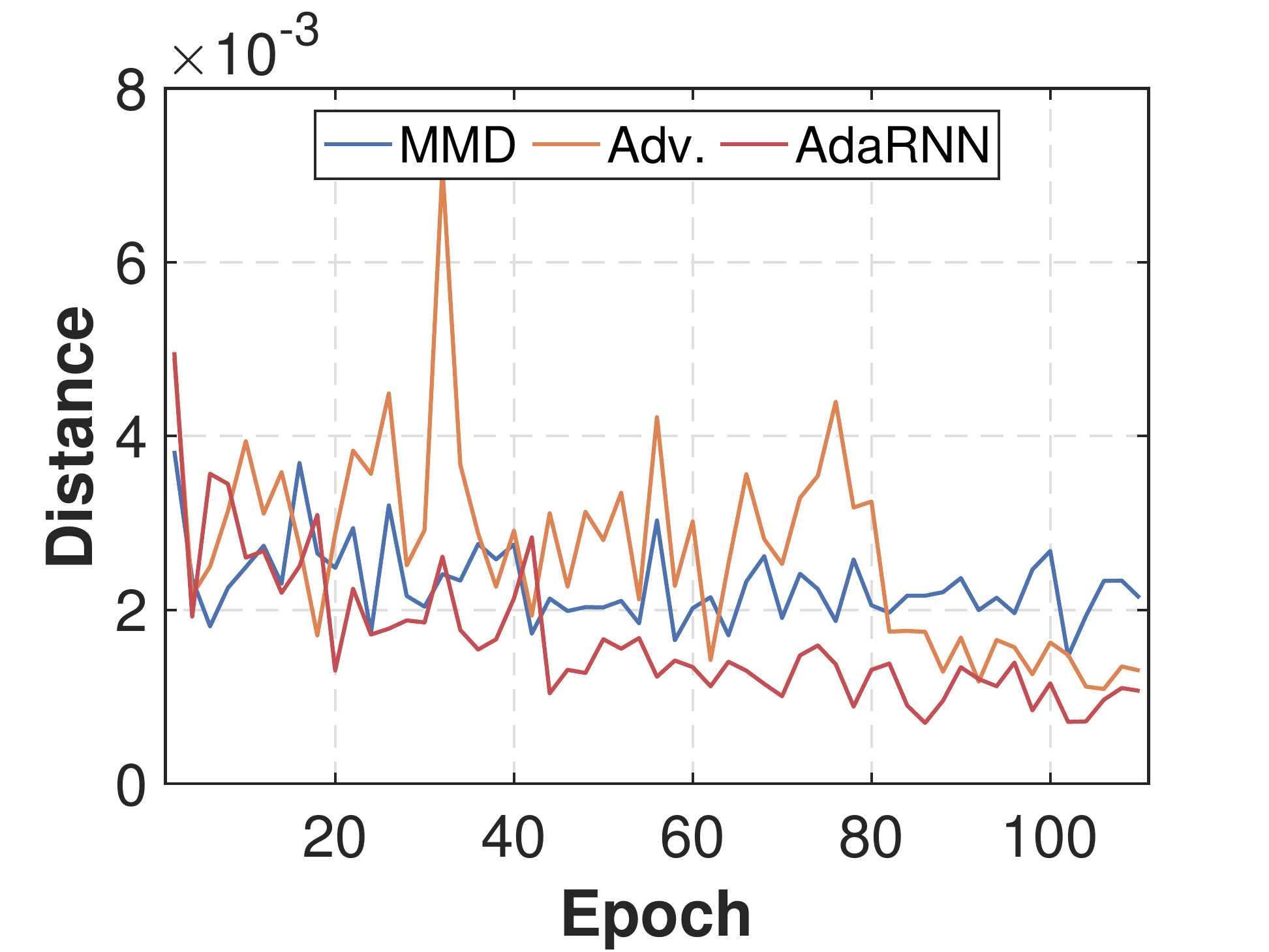}
		\label{fig-sub-mmd}}
	\subfigure[Multi-step prediction]{
		\centering
		\includegraphics[width=.19\textwidth]{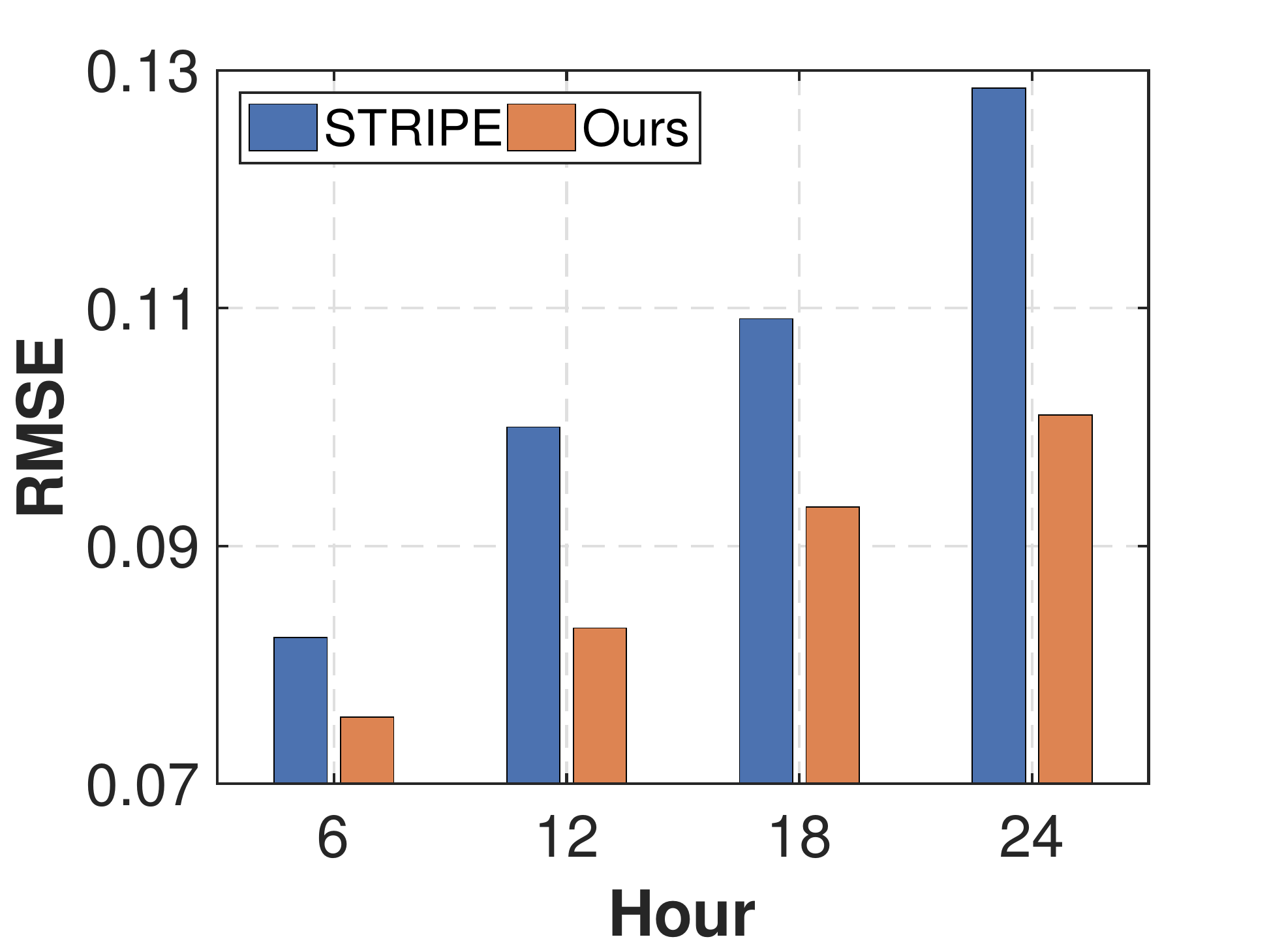}
		\label{fig-sub-multistep}}
	\subfigure[Convergence]{
		\centering
		\includegraphics[width=.19\linewidth]{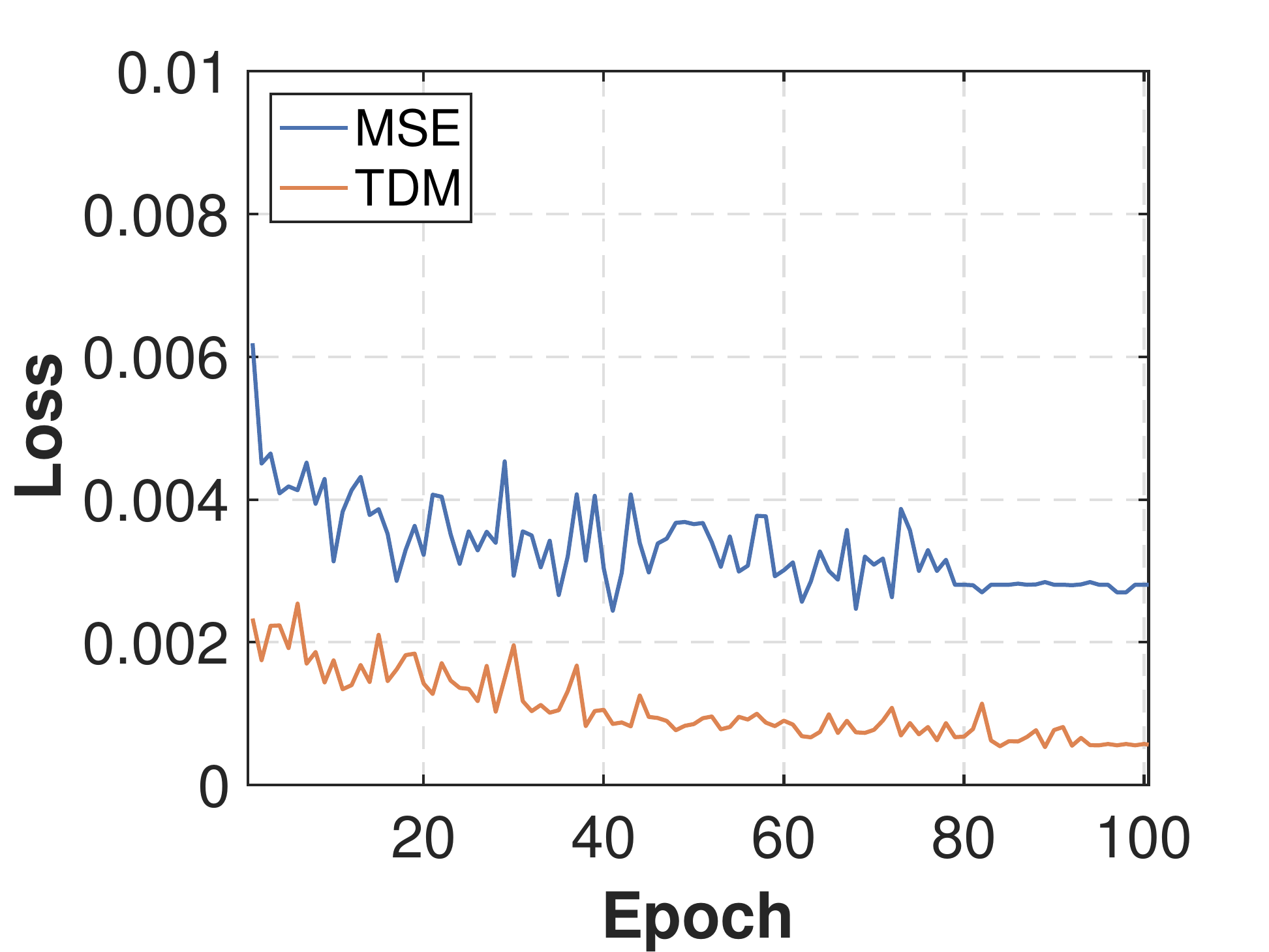}
		\label{fig-sub-conv}}
	\subfigure[Training time]{
		\centering
		\includegraphics[width=.19\linewidth]{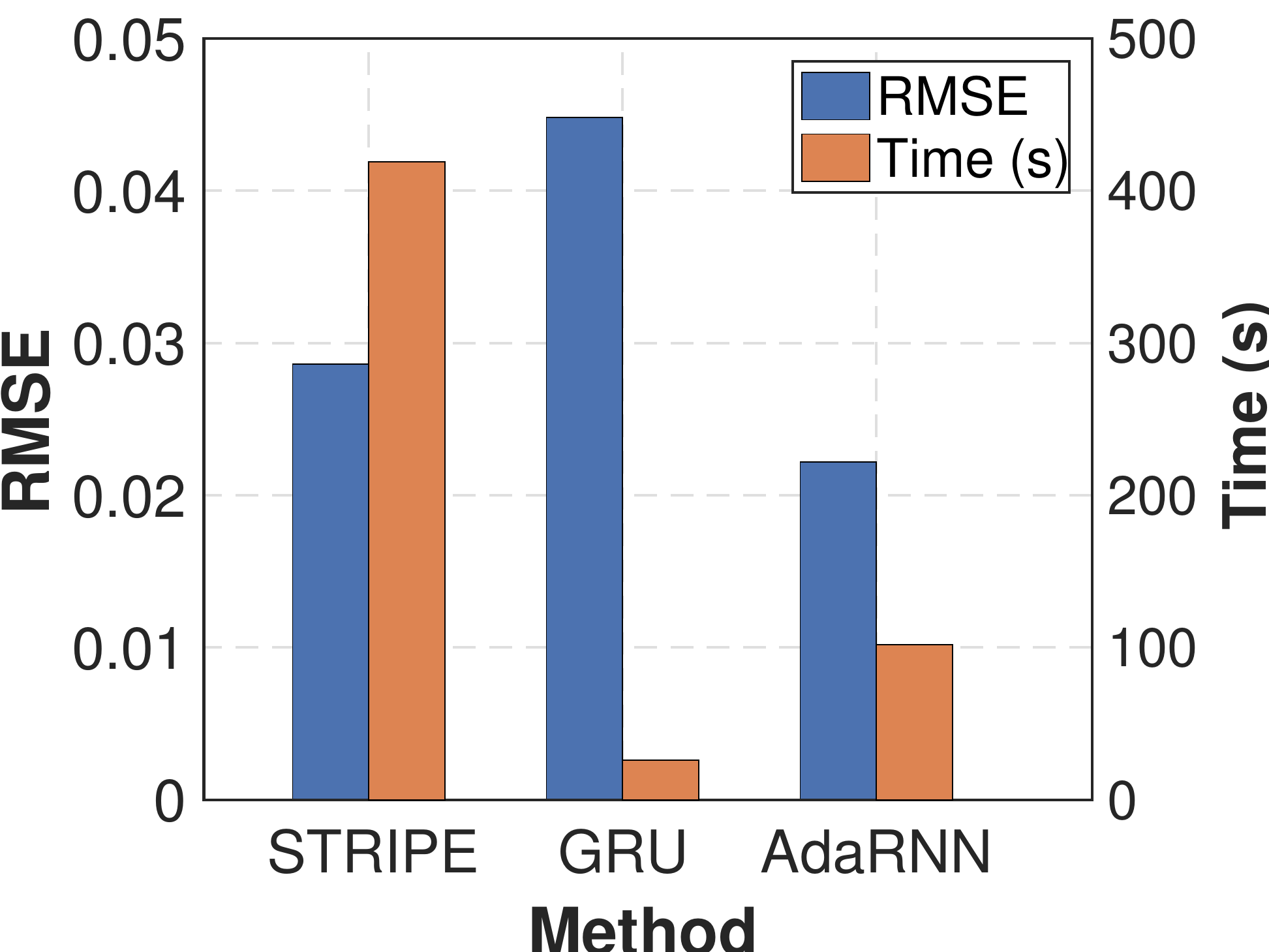}
		\label{fig-sub-time}}
	\vspace{-.1in}
	\caption{(a) Multi-step prediction of air quality. (b) TDM is agnostic to distribution distances in different stations. (c) The change of distribution distance. (d) Convergence analysis. (e) Training time.}
	\label{fig-analysis}
	\vspace{-.1in}
\end{figure*}

\subsubsection{Temporal distribution matching}

We analyze the TDM module by evaluating the performance of pre-training and the boosting-based method. We compare four variants of TDM: (1) na\"ive approach (i.e., learning $\bm{\alpha}$ by a feed-forward layer as discussed in remark \ref{remark}) w./o pre-training, (2) boosting w./o pre-training (3) pre-training + na\"ive, (4) pre-training + boosting (Ours).

\figurename{~\ref{fig-sub-tdm}} shows the performance on two stations. 
We observe that the pre-training technique helps both na\"ive and boosting-based methods to get better performance. 
More importantly, our boosting-based method can achieve the best performance with pre-training. 
Compared with the STRIPE method from \tablename{~\ref{tb-air-quality}}, we see that almost all four variants of TDM can achieve better performance than STRIPE. 
We also observe that by manually setting the same importance to each hidden state, the results are not comparable. 
To summarize, it is necessary to consider the importance of each hidden state when trying to build an adaptive RNN model.

To evaluate the importance vector $\bm{\alpha}$, we compare the results of with or without importance vector in \figurename~\ref{fig-sub-alpha}.
The results indicate that by taking the same distribution distance, our \method achieve the best results.
This demonstrates that the different distributions of each hidden state is important to model in RNN-based time series modeling, 
to which our AdaRNN is capable of dynamically evaluating their importance to achieve the best performance.

\subsection{Further analysis}

\subsubsection{Distribution distance}
\method is agnostic to the distribution distance. Here we evaluate this property by running TDM with four different kinds of distribution distances: MMD~\cite{borgwardt2006integrating}, CORAL~\cite{Sun2016DeepCC}, Cosine, and domain-adversarial distance~\cite{Ganin2016DomainAdversarialTO}. \figurename~\ref{fig-sub-distance} presents the results on three stations for air quality prediction. Combining the results from \tablename~\ref{tb-air-quality}, we observe that \method is robust to the choice of distribution distance while achieving the best performance.
Moreover, \figurename~\ref{fig-sub-mmd} shows that by adding the importance weights $\bm{\alpha}$, our \method achieves lowest distribution distances compared to MMD and adversarial training.

\subsubsection{Multi-step prediction}
We analyze the performance of \method for multi-step ($r > 1$) air quality prediction and compare with STRIPE~\cite{leguen20stripe} by predicting air quality index in next $6,12,18,24$ hours (the output sequence length is $6,12,18,24$ accordingly). \figurename{~\ref{fig-sub-multistep}} shows the results on one station.
We see that \method achieves the best performance for multi-step prediction. 
In addition, as the prediction steps increase ($6 \rightarrow 12 \rightarrow 18 \rightarrow 24$), 
the performances of all methods are becoming worse 
since it is more difficult to model the longer sequences and also affected by the cumulative errors.

\subsubsection{Convergence and time complexity analysis}
We analyze the convergence of \method by recording its training loss for prediction and distribution matching in \figurename~\ref{fig-sub-conv}. We see that \method can be easily trained with quick convergence.
On the other hand, \method is based on RNNs which are autoregressive models. So it is reasonable that \method takes more time for training than traditional RNNs. \figurename~\ref{fig-sub-time} shows the training time comparison between \method, GRU, and STRIPE~\cite{leguen20stripe}. We see that other than the best performance, the training time of \method is still comparable to existing methods. The inference time of \method is the same as existing RNN-based models since we only need to compute Eq.~\eqref{eq-mse}.

\begin{figure}[t!]
	\centering
	\vspace{-.1in}
	\subfigure[MMD-RNN]{
		\centering
		\includegraphics[width=.46\linewidth]{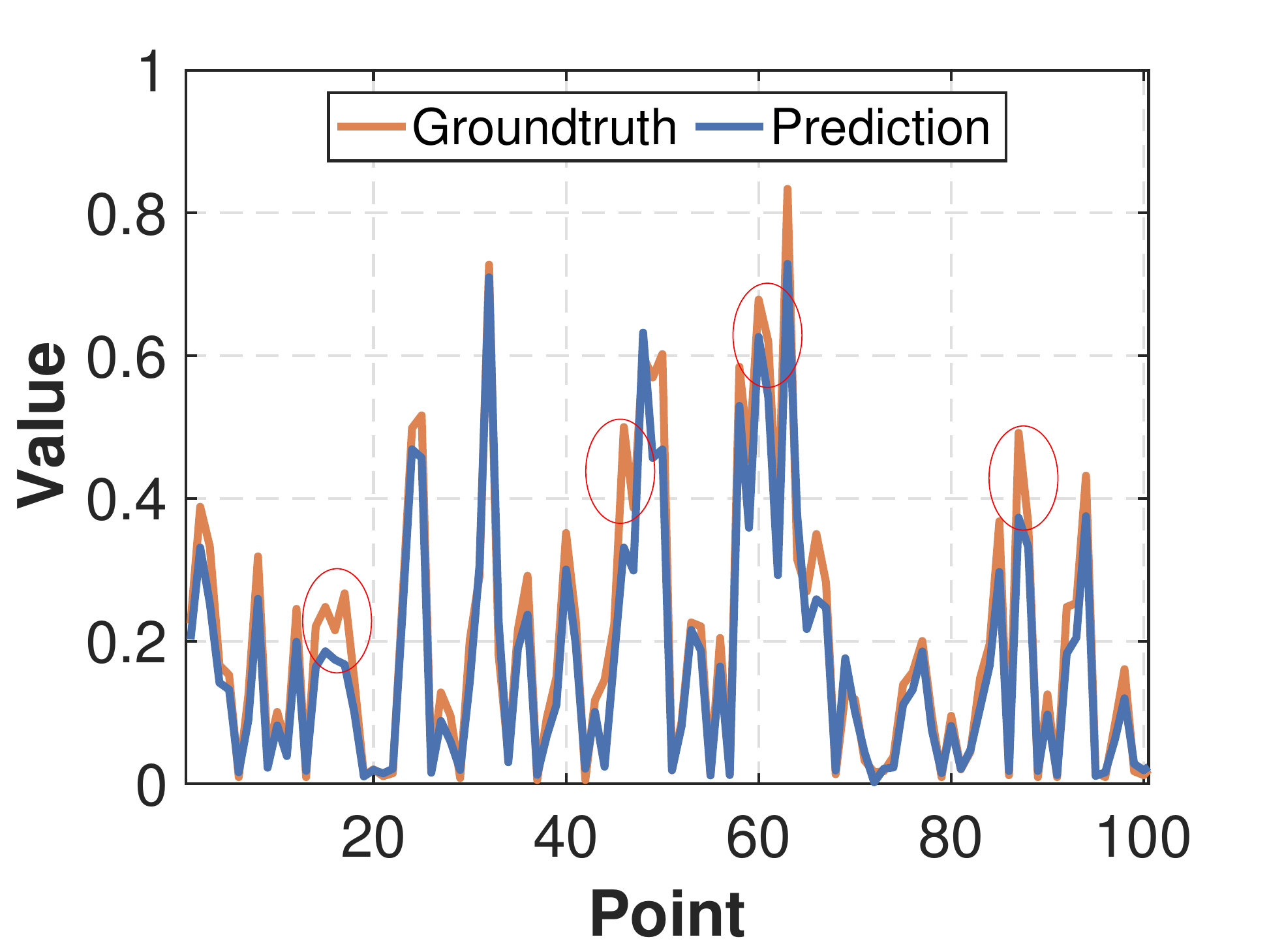}
		\label{fig-pred-mmd}}
	\subfigure[DANN-RNN]{
		\centering
		\includegraphics[width=.46\linewidth]{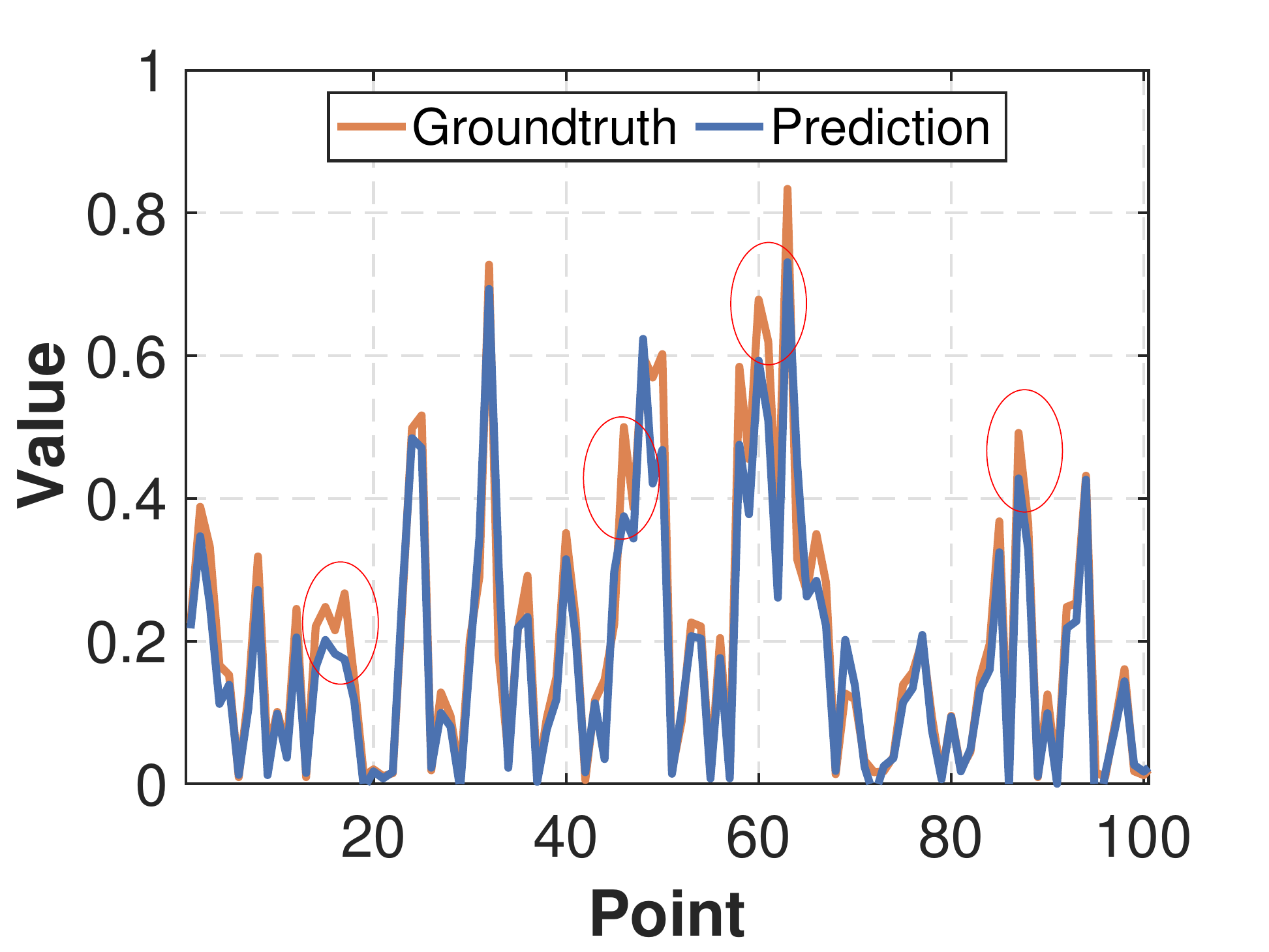}
		\label{fig-pred-dann}}
	\\ \,\, 
	\subfigure[\method]{
		\centering
		\hspace{-.2in}
		\includegraphics[width=.46\linewidth]{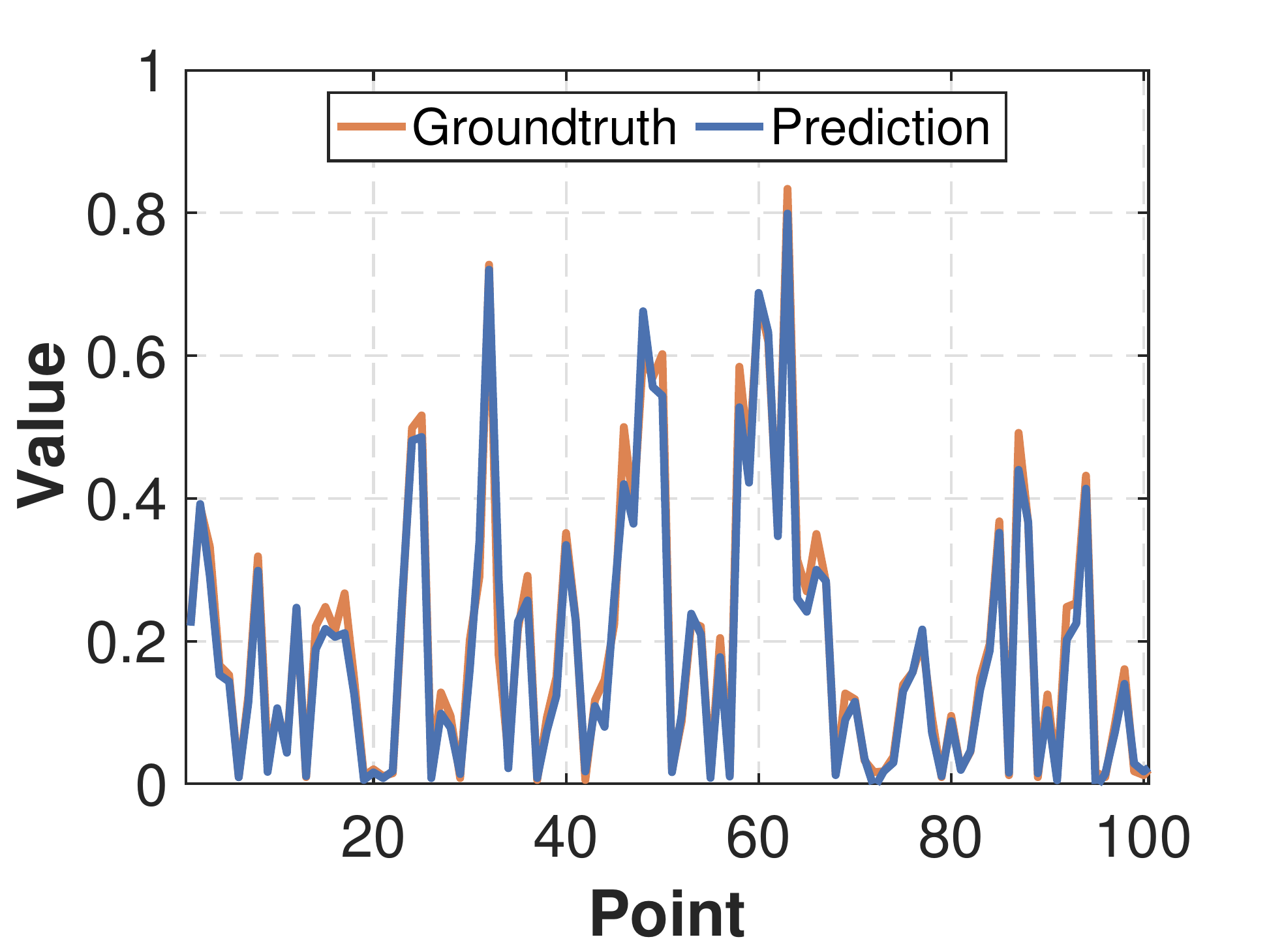}
		\label{fig-pred-adarnn}}
	\subfigure[Differences]{
		\centering
		\includegraphics[width=.46\linewidth]{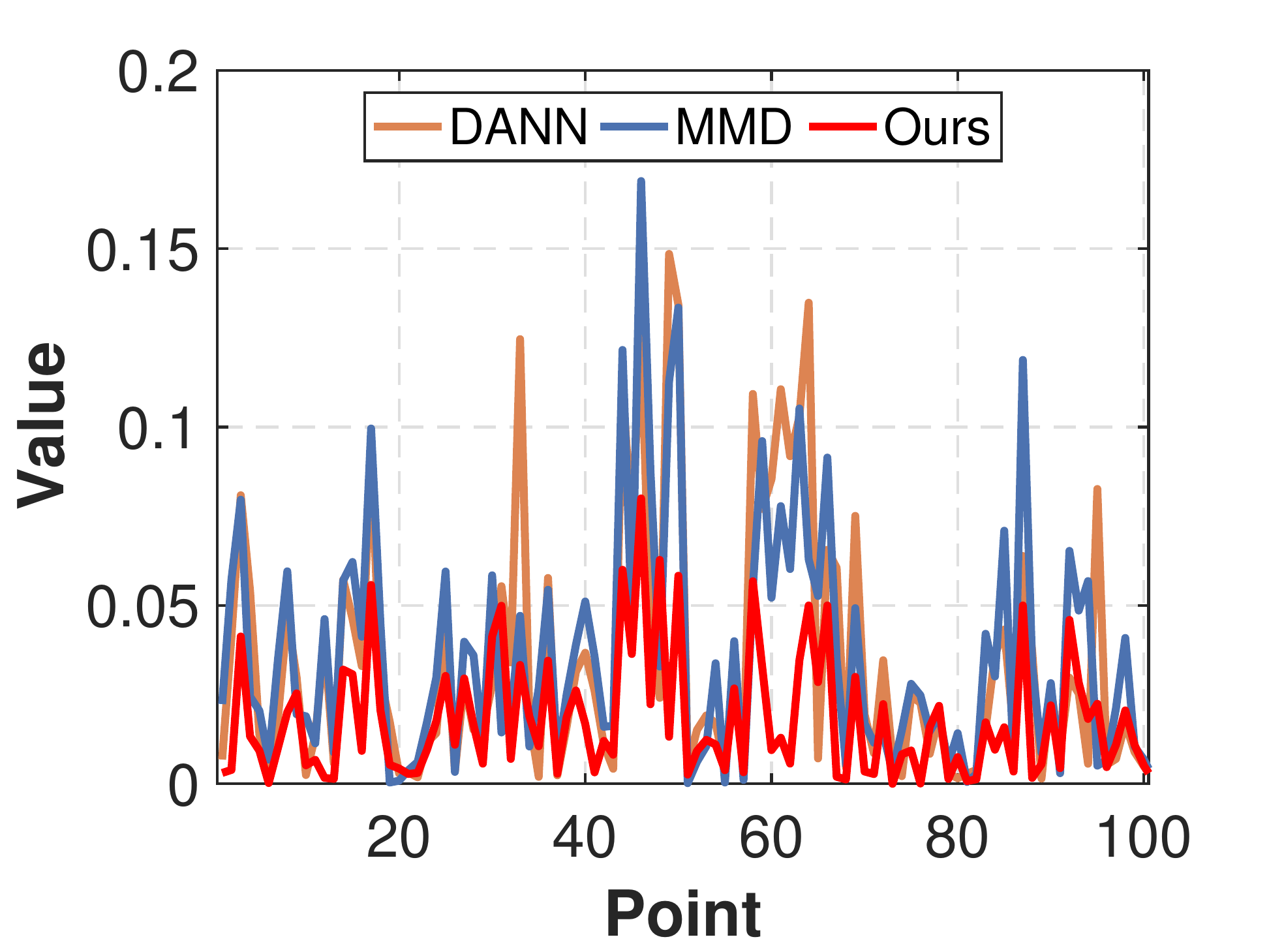}
		\label{fig-pred-compare}}
	\vspace{-.1in}
	\caption{Prediction vs. ground truth.}
	\label{fig-pred}
	\vspace{-.1in}
\end{figure}

\subsection{Case study: prediction vs. ground truth}
Taking the station Dongsi as an example, we show the prediction vs. ground truth values of different methods in \figurename~\ref{fig-pred}, where the red circles denote the bad cases.
We see that DANN and MMD achieve worse results than our \method since they have relatively worse predictions in many points.
Especially in \figurename~\ref{fig-pred-compare}, we show the absolute difference between the prediction of each method and the ground-truth.
We can clearly see that our \method outperforms all methods with much less mistakes.

\section{Extensions}

Apart from RNN-based models, in this section, we add the adaptive importance weights to each hidden representations in the Transformer~\cite{vaswani2017attention} model, which is claimed more powerful than RNN-based models in the latest studies.
We add the importance weights $\bm{\alpha}$ to the representations of each self-attention block and term this extension as \emph{AdaTransformer}.
As shown in \tablename~\ref{tb-adatransformer}, this pilot study shows the power of the \method in Transformers.
We believe more improvements can be made in the future.

\begin{table}[htbp]
    \vspace{-.1in}
    \caption{Results of vanilla and adaptive transformer}
    \vspace{-.1in}
    \label{tb-adatransformer}
    \begin{tabular}{lcc}
        \toprule
        Method & Station 1 & Station 2 \\ \hline
        Vanilla Transformer & 0.028 & 0.035 \\ 
        AdaTransformer & \textbf{0.025} & \textbf{0.030} \\ \bottomrule
    \end{tabular}
    \vspace{-.1in}
\end{table}

\section{Conclusions and Future Work}
This paper proposed to study the more realistic and challenging temporal covariate shift (TCS) problem for non-stationary time series. We proposed the \method framework to learn an adaptive RNN model for better generalization. \method consists of two novel algorithms: \tsqfull to characterize the distribution information in TS and temporal distribution matching to build the generalized RNN model by distribution matching. Extensive experiments show its effectiveness in three datasets.

In the future, we plan to extend this work from both algorithm and application level. First, explore deeper extension of \method to Transformer for better performance. Second, integrate the two stages of \method in an end-to-end network for easier training.

\appendix

\section{Distribution Distance}

We adopt four widely used distribution distance functions following existing work on DA/DG, i.e., cosine, MMD~\cite{borgwardt2006integrating}, CORAL~\cite{Sun2016DeepCC}, and domain adversarial discrepancy~\cite{Ganin2016DomainAdversarialTO}.

\textbf{Cosine}. Cosine distance is widely used in time series data~\cite{Wang2015ImagingTT,Lei2019SimilarityPR} to measure the similarity between two distributions. Given the data $\mathbf{h}_s$ and $\mathbf{h}_t$, cosine distance is defined as:
\begin{equation}
d_{cosine}(\mathbf{h}_s,\mathbf{h}_t) = 1 - \frac{\left<\mathbf{h}_s, \mathbf{h}_t\right>}{||\mathbf{h}_s||\cdot||\mathbf{h}_t||}.
\end{equation}

\textbf{Maximum Mean Discrepancy (MMD)}. The MMD \cite{Gretton2012OptimalKC} is a non-parametric distance between two probability distributions. MMD measures the distance of the mean embeddings of the samples from two distributions in a Reproducing Kernel Hilbert Space (RKHS). MMD can be empirically estimated by
\begin{small}
	\begin{equation}
	\begin{aligned}
	d_{mmd}(\mathbf{h}_s, \mathbf{h}_t) &= \frac{1}{n_s^2}\sum_{i,j=1}^{n_s}k(h_{s_i}, h_{s_j}) + \frac{1}{n_t^2}\sum_{i,j=n_s + 1}^{n_s + n_t}k(h_{s_i}, h_{s_j}) \\
	& - \frac{2}{n_sn_t}\sum_{j=n_s+1}^{n_s+n_t}k(h_{s_i}, h_{s_j}),
	\end{aligned}
	\end{equation}
\end{small}
where $k(\cdot,\cdot)$ is the kernel function such as RBF kernel and linear kernel, and $n_s = |h_s|, n_t = |h_t|$ are the number of the data from two distributions, respectively.

\textbf{Deep Correlation Alignment (CORAL)}. The CORAL distance \cite{Sun2016DeepCC} is defined as the distance of the second-order statistic (covariance) of the samples from two distributions:
\begin{equation}
d_{coral}(\mathbf{h}_s, \mathbf{h}_t) = \frac{1}{4q^2}||\mathbf{C}_s - \mathbf{C}_t||_F^2,
\end{equation}
where $q$ is the dimension of the features, and $\mathbf{C}_s$ and $\mathbf{C}_t$ are the covariance matrices of the two distributions, respectively.

\textbf{Domain Adversarial Discrepancy}. The domain discrepancy can be parameterized as a neural network and it is referred to as the domain adversarial discrepancy \cite{Ganin2016DomainAdversarialTO}. An additional network named domain discriminator denoted by $D$ is introduced. The domain adversarial objective is defined as:
\begin{equation}
\begin{aligned}
l_{adv}(\mathbf{h}_s, \mathbf{h}_t) = \mathbb{E}[\log[D(\mathbf{h}_s)]] + \mathbb{E}[\log[1 - D(\mathbf{h}_t)]],
\end{aligned}
\end{equation}
where $\mathbb{E}$ denotes expectation. Hence, the adversarial discrepancy is
\begin{equation}
d_{adv}(\mathbf{h}_s,\mathbf{h}_t) = -l_{adv}(\mathbf{h}_s,\mathbf{h}_t).
\end{equation}

\section{Different distances in TDC}


We analyze temporal distribution characterization (TDC) by using different distance functions. We set $K = 2$ for TDC and compare its performance with one random domain split and the opposite domain split (i.e., with the smaller distribution distance).

The results in \tablename~\ref{tb-DD} indicate that our TDC algorithm achieves the best performance in RMSE and MAE with the largest distribution distance under most distance functions. When using cosine distance, our split achieves the second largest distribution distance but it gets the best performance in RMSE and MAE. The results show
that TDC could effectively characterize the distribution information and is agnostic to distance functions.

\begin{table}[tbp]
	\centering
	\caption{Results of different distances in TDC}
	\label{tb-DD}
	\vspace{-.1in}
	\resizebox{.5\textwidth}{!}{
	\begin{tabular}{cccccccc}
		\toprule
	    Split	&  $|\mathcal{D}_1|/|\mathcal{D}_2|$     & $d$(MMD)        & $d$(DTW)  & $d$(CORAL) &    $d$(Cosine)  & RMSE            & MAE             \\
		\hline
		Split1 & 0.75/0.25 & 0.898 & 2.54   &  9.9999884E-1 &  0.9748   & 0.0233  & 0.0179          \\
		Split2 & 0.25/0.75 & 0.910 & 2.60 &  9.9999902E-1 &   \textbf{0.9966}    & 0.0239  & 0.0197          \\
		Ours    & 0.66/0.33 & \textbf{0.932} & \textbf{2.79}  & \textbf{9.9999919E-1} & 0.9861 & \textbf{0.0164} & \textbf{0.0112} \\
		\bottomrule
	\end{tabular}
	}
	\vspace{-.1in}
\end{table}

\section{Stock price prediction: Metrics}

The evaluation metrics for stock price prediction have been widely adopted in previous work~\cite{kohara1997stock}.
The information coefficient (IC) describes the correlation between predicted and actual stock returns, sometimes used to measure the contribution of a financial analyst. The IC can range from $1.0$ to $-1.0$. An IC of $+1.0$ indicates a perfect linear relationship between predicted and actual returns, while an IC of $0.0$ indicates no linear relationship. An IC of $-1.0$ indicates that the analyst always fails at making a correct prediction. IC is defined as:
\begin{equation}
IC = corr(f_{t-1} , r_t),
\end{equation}
where $f_{t-1}$ is the factor value of the stock in period $t-1$ and $r_t$ is the stock return rate of period $t$. $corr(\cdot,\cdot)$ is the  correlation functions. Besides, there are also some variants of IC, such as rankIC, which is
\begin{equation}
rankIC = corr({order}_{t-1}^f , {order}_t^r),
\end{equation}
where ${order}_{t-1}^f$ is the factor rank of each stock in period $t-1$ and ${order}_t^r$ is the ranking of each stock's return rate in period $t$.

The information ratio (IR) of an actively managed portfolio is defined as the portfolio's mean return over a benchmark portfolio return divided by the tracking error (standard deviation of the excess return), which is defined as
\begin{equation}
IR = IC \times \sqrt{BR},
\end{equation}
where BR is breadth or number of independent information sources.

We also call IR the ICIR. Based on IC and ICIR, we further use RankIC and RankICIR to evaluate the robustness of models against permuations, and use RawIC and RawICIR to denote the performance of unnormalized data. To sum up, the higher they are, the better the model is.

	

	


\bibliographystyle{ACM-Reference-Format}
\bibliography{refs}


\end{document}